\documentclass[10pt,twocolumn,letterpaper]{article}

\usepackage[pagenumbers]{wacv} % To force page numbers, e.g. for an arXiv version

\newcommand{\RNum}[1]{\lowercase\expandafter{\romannumeral #1\relax}}
\newcommand{\RNumUp}[1]{\uppercase\expandafter{\romannumeral #1\relax}}

\renewcommand{\etal}{et~al.}
\newcommand{\methodname}{ChromaDistill}

\usepackage{graphicx}
\usepackage{amsmath}
\usepackage{amssymb}
\usepackage{booktabs}

\usepackage[ruled,vlined]{algorithm2e}
\usepackage[noend]{algpseudocode}
\usepackage{wrapfig}
\usepackage{adjustbox}

\usepackage[pagebackref,breaklinks,colorlinks]{hyperref}

\usepackage[capitalize]{cleveref}
\crefname{section}{Sec.}{Secs.}
\Crefname{section}{Section}{Sections}
\Crefname{table}{Table}{Tables}
\crefname{table}{Tab.}{Tabs.}

\def\wacvPaperID{435} % *** Enter the WACV Paper ID here
\def\confName{WACV}
\def\confYear{2025}

\begin{document}

\title{ChromaDistill: Colorizing Monochrome Radiance Fields with Knowledge Distillation}

% \and

\author{
Ankit Dhiman\textsuperscript{1,2} \quad
R Srinath\textsuperscript{1} \quad
Srinjay Sarkar\textsuperscript{1} \quad
Lokesh R Boregowda\textsuperscript{2} \quad 
R Venkatesh Babu\textsuperscript{1} \\
\textsuperscript{1}Vision and AI Lab, IISc Bangalore \quad
\textsuperscript{2}Samsung R \& D Institute India - Bangalore \quad
}

\maketitle

\begin{figure*}
    \centering
    \includegraphics[width=0.95\textwidth]{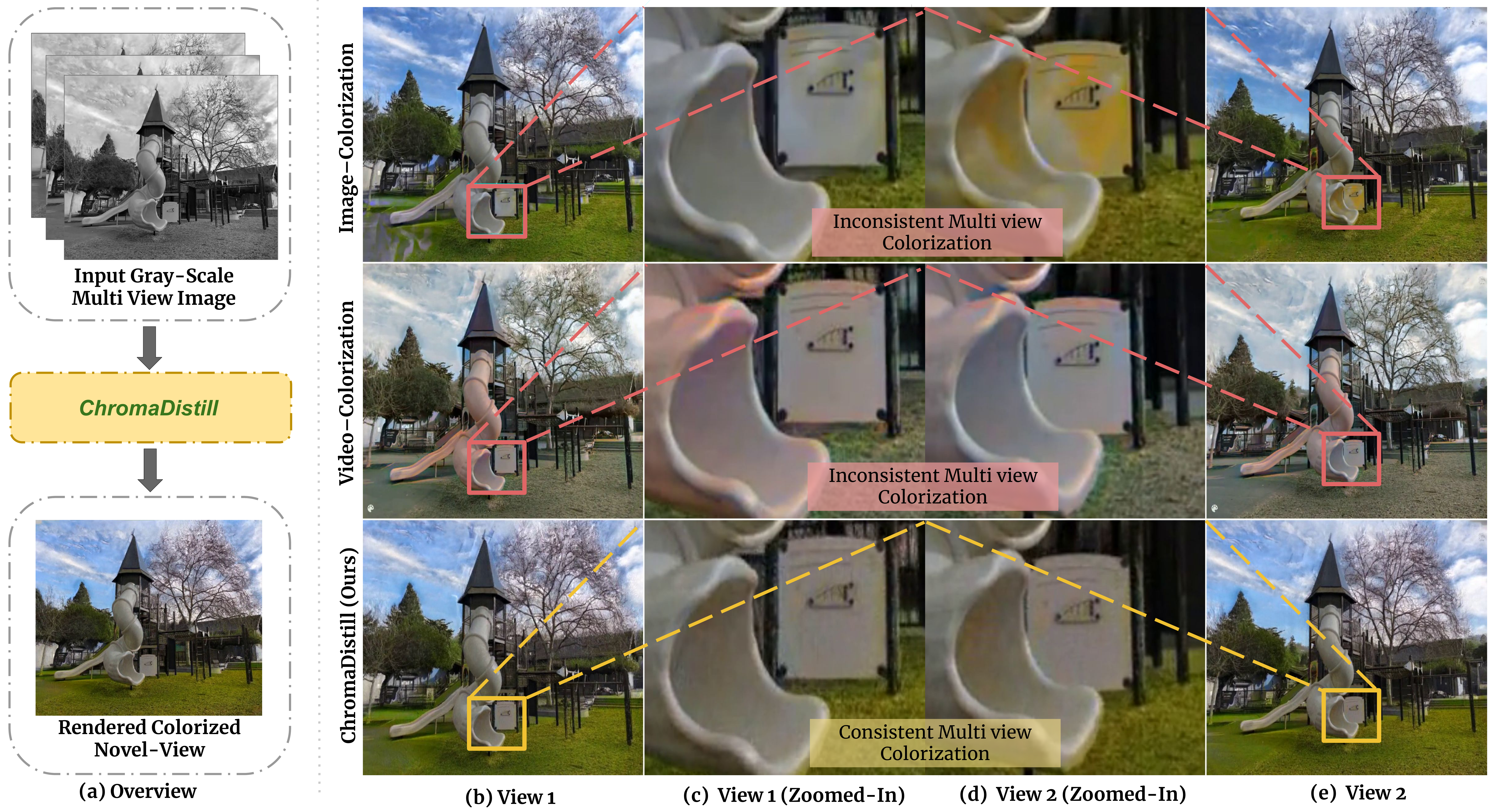}
    \vspace{-2mm} 
  \caption{(a) Overview of our method. Given input multi-view gray-scale views, the proposed approach ``\methodname{}'' is able to generate colorized views which are 3D consistent. Two colorized novel-views (b) and (e) by \RNumUp{1}. Image-colorization baseline, \RNumUp{2}. Video-colorization baseline, and \RNumUp{3}. our approach on ``playground'' scene from LLFF~\cite{mildenhall2019llff} dataset. State-of-the-art colorization baselines generate 3D inconsistent novel-views as shown in zoomed-in regions in (c) and (d).}
  \label{fig:teaser_image}
  \vspace{-2mm}
\end{figure*}  

\begin{abstract}
Colorization is a well-explored problem in the domains of image and video processing. However, extending colorization to 3D scenes presents significant challenges. 
%Traditional 3D representations primarily rely on geometric cues, which are insufficient for effectively performing the colorization task. 
Recent Neural Radiance Field (NeRF) and Gaussian-Splatting (3DGS)  methods enable high-quality novel-view synthesis for multi-view images. 
%These representations are suitable for colorization as the loss is propagated through the rendered images. 
%Beyond its visual aesthetic significance in portraying worlds, colorizing monochromatic signals is crucial for downstream applications such as open-set scene decomposition etc. 
However, the question arises: \textit{How can we colorize these 3D representations?} This work presents a method for synthesizing colorized novel views from input grayscale multi-view images. Using image or video colorization methods to colorize novel views from these 3D representations naively will yield output with severe inconsistencies. 
%Another approach is to train the 3D representation by colorizing the grayscale training views. However, due to output inconsistencies from image or video colorization techniques, training a 3D representation with these colorized images results in artifacts in the novel views. 
We introduce a novel method to use powerful image colorization models for colorizing 3D representations. We propose a distillation-based method that transfers color from these networks trained on natural images to the target 3D representation. Notably, this strategy does not add any additional weights or computational overhead to the original representation during inference. Extensive experiments demonstrate that our method produces high-quality colorized views for indoor and outdoor scenes, showcasing significant cross-view consistency advantages over baseline approaches. Our method is agnostic to the underlying 3D representation and easily generalizable to NeRF and 3DGS methods. Further, we validate the efficacy of our approach in several diverse applications: 1.) Infra-Red (IR) multi-view images and 2.) Legacy grayscale multi-view image sequences. 
Project Webpage:  \href{https://val.cds.iisc.ac.in/chroma-distill.github.io/}{https://val.cds.iisc.ac.in/chroma-distill.github.io/}

\end{abstract}
\vspace{-4mm} 

\addtocontents{toc}{\protect\setcounter{tocdepth}{-2}}
\section{Introduction}
\noindent
Adding color to a monochromatic signal is a longstanding problem~\cite{larsson2016learning,iizuka2016let, cheng2015deep, larsson2016learning,zhang2016colorful} in computer vision and graphics. This monochromatic signal can be obtained from special sensors such as infra-red (IR) sensors or legacy content such as old movies. A range of methods have been proposed to colorize images/videos~\cite{lai2018learning, meyer2018deep,vondrick2018tracking}; however, colorization of 3D scenes is challenging as it needs to maintain 3D consistency for realistic colorization. Recent exploration of 3D representations (e.g., NeRF~\cite{mildenhall2021nerf} and 3DGS~\cite{kerbl3Dgaussians}) has enabled effective modeling of complex real-world 3D scenes given multi-view images. Leveraging this, we formulate the problem of colorization of 3D representations given input multi-view grayscale images of a scene. To solve this effectively, we raise the following question: \textit{Can we leverage rich knowledge learned from existing image colorization approaches to colorize these 3D representations?}

% colorizing these 3D representations such as NeRF or hybrid representations such as Gaussian Splatting~\cite{kerbl3Dgaussians}(3DGS) in a 3D consistent manner from monochromatic input multi-view images to generate colorized novel-views. 
\noindent
This practical setting for colorizing 3D representations has many applications: \textbf{\textit{a)}} generating colorized novel views from legacy images/videos, \textbf{\textit{b)}} generating colorized novel views from monochromatic signals such as IR and \textbf{\textit{c)}} enhancing the performance of discriminative models (e.g., object detection)~\cite{xu2023nerf} on monochromatic signals by applying colorization prior to inference. A straightforward approach to colorize a 3D representation involves applying image colorization methods~\cite{kim2022bigcolor, zhang2016colorful} to the input views before training the 3D representation. However, this simplistic method leads to 3D inconsistencies (Fig.~\ref{fig:teaser_image}) across views since each view is independently colorized, resulting in inconsistent color assignments to the same 3D point. Another promising approach is to use video colorization methods on the generated novel-view sequence. This approach ensures temporal consistency (Fig.~\ref{fig:teaser_image}) but fails to guarantee 3D consistent colorization, as it is not grounded in 3D representation.
% Due to its impact on downstream tasks, recent research has focused on solving the colorization problem for images~\cite{cong2024automatic,zabari2023diffusing} and videos~\cite{chen2024exemplar, yang2024bistnet}. 

% We can apply image colorization methods to the input grey-scale images and train a radiance field network, but the generated novel views will not be 3D consistent. 

% The colorization task is also extended to video domain~\cite{lai2018learning, meyer2018deep,vondrick2018tracking}. 

% However, it is not well addressed for effective colorization of 3D scenes. This is especially challenging as 3D representations. The key challenge to adapting this method for the colorization of 3D representation is that this representation only has geometric hints, unlike the image domain, which has a strong grey-scale hint. 
% Previously, there exist works~\cite{shinohara2021point2color, Wu_2023_ICCV} which colorize a 3D point-cloud but they are only successful for simple shapes.  

% Recently proposed approaches GSN~\cite{devries2021unconstrained}, GRAF~\cite{schwarz2020graf} aim to directly colorize 3D representations by 
\noindent
In the context of 3D colorization, earlier works~\cite{shinohara2021point2color, Wu_2023_ICCV} tried to colorize point clouds. Another direction is to add texture to a predefined mesh~\cite{bokhovkin2023mesh2tex, yu2023texture}. However, these methods are limited to simple synthetic objects. 
%or need manual processing to generate refined meshes from real-world captures. 
Recent 3D representations (e.g., NeRF, 3DGS) effectively capture high-quality geometry of real-world scenes and can propagate losses from 2D images due to their differentiable nature. 
This underscores a need for novel techniques to colorize these representations and enable realistic colorization of complex real scenes.

\noindent
To colorize these 3D representations, a recent method ~\cite{wang2024lift3d} lifts the encoder features of any 2D vision model to the 3D representation. The features are rendered in 2D and passed through the vision model's decoder to generate a consistent novel view and obtain the final RGB image. However, the features are encoded at a very low resolution and may lead to inconsistent 3D colorization due to independent decoding with the image space decoder. We hypothesize that there are two crucial requirements for high-quality 3D colorization: \textbf{\textit{i)}} 3D consistency and \textbf{\textit{ii)}} accurate colorization with minimal bleeding artifacts.

%Colorization holds tremendous potential, particularly AR/VR applications, especially in restoring legacy content. %\href{https://www.reddit.com/r/GaussianSplatting/comments/1c8w3h1/going_into_iconic_movie_scenes_using_gaussian/}{legacy content}. It is also useful for modalities that lack color but capture detailed structure in the scene, such as NIR and monochromatic cameras. These sensors have applications in 3D sensing, low-light enhancement, and AR/VR headsets. While NIR image colorization~\cite{suarez2018learning, suarez2017infrared, suarez2017colorizing} has notable applications in surveillance and remote sensing, the exploration of consistent multi-view colorization remains limited. %Mainstream large-scale pre-trained models are trained for color images and do not realize their full potential when directly applied to modalities such as IR images. Consequently, addressing the gap in multi-view colorization is crucial to enhance downstream applications for advanced vision tasks.
\noindent
In this work, we propose a novel framework for accurate colorization of 3D representations by distilling knowledge from state-of-the-art image colorization methods. We introduce a two-stage process for effective colorization. In the first stage, we train a luminance radiance field to learn the geometry from grayscale images effectively. In the second stage, we freeze the geometry and distill the chroma component from the pre-trained image colorization network. This two-stage training effectively decouples the geometry and colorization of the 3D scenes, leading to high-quality colorization outputs with refined geometry.
Notably, this strategy incurs no additional cost for training a separate colorization module for the radiance field networks. Further, we propose a novel \textit{multi-scale self-regularization} technique to mitigate the desaturation (washed-out color) effects when distilling from the colorization network.

\noindent
We demonstrate the effectiveness of our approach in colorizing both front-facing and unbounded 3D scenes from widely used 3D datasets ~\cite{mildenhall2019llff,knapitsch2017tanks,shiny}. Our method significantly outperforms all the baselines in multi-view consistency and realism of colorization. We also compare favorably to state-of-the-art stylization approaches. Further, we show results on two downstream tasks: \textbf{1)} Colorizing multi-view infrared (IR) images and \textbf{2)} Colorizing legacy grayscale content.
Notably, when used for the downstream object detection task, the colorized IR images significantly improve the detection scores. Our primary contributions are:

% We show the effectiveness of our approach on various grayscale image sequences generated from widely used datasets such as: \  LLFF~\cite{mildenhall2019llff}, Tanks \& Temples~\cite{knapitsch2017tanks} and Shiny~\cite{shiny}. The scenes we examined vary in complexity. Our experiments demonstrate that our approach is effective for both forward-facing and unbounded scenes. Detailed examples with videos are provided in the supplementary material. Additionally, we quantitatively show that our method achieves superior multi-view consistency, outperforming all baseline methods in this metric. Furthermore, we establish that existing stylization methods for 3D representations are inadequate for the colorization task. Further, we show results on two downstream tasks: 1.) Colorizing multi-view infrared (IR) images and 2.) Colorizing In-the-wild grayscale content. Additionally, we demonstrate the effectiveness of our approach for downstream tasks, such as object detection, after colorizing the infrared (IR) views. Our results show that objects are detected more accurately in the colorized views compared to the original monochromatic IR views. Our main contributions are:

\begin{itemize}
    \vspace{-2mm}
     \item We introduce a novel approach \textit{\methodname{}} for colorizing radiance field networks to produce 3D consistent colorized novel views from input grayscale multi-view images. 
    \item We propose a multi-scale self-regularization to mitigate de-saturation in the distilled color.
    \item We show that the proposed colorization approach is generalizable to any 3D representation e.g 3DGS~\cite{kerbl3Dgaussians} and NeRF~\cite{mildenhall2021nerf}.
    \item We demonstrate our approach on two real-world applications for novel view synthesis: input multi-view IR images and input grayscale legacy content.
\end{itemize}
%\vspace{-0.5em}
\begin{figure*}[!t]
    \centering
    \includegraphics[width=0.90\linewidth]{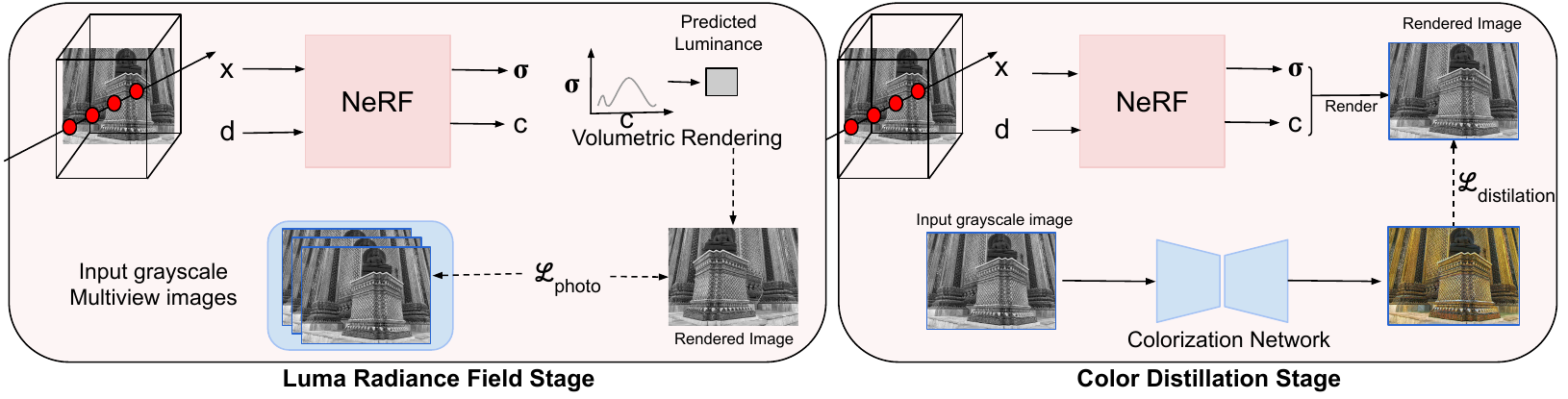}
    \caption{Overall architecture of our method. First, we train a radiance field network from input multi-view grayscale images in the ``Luma Radiance Field Stage''. Next, we distill knowledge from a teacher colorization network trained on natural images to the radiance field network trained in the previous stage.}
    \label{fig:method}
    \vspace{-1em}
\end{figure*}
\section{Related Work}
\noindent
\textbf{Image Colorization.} One of the earliest deep-learning based methods~\cite{iizuka2016let} used a CNN to estimate color for the grayscale images by jointly learning global and local features. Larsson~\etal~\cite{larsson2016learning} train the model to predict per-pixel color histograms by leveraging pre-trained networks for high and low-level semantics. Zhang~\etal~\cite{zhang2017real}  also colorize a grayscale image using a CNN network. GANs have also been used for the image colorization task.~\cite{vitoria2020chromagan} uses a generator to produce the chroma component of an image from a given grayscale image, which is conditioned on semantic cues. GAN methods exhibit strong generalization to new images. Recently, diffusion-based methods~\cite{zabari2023diffusing, cong2024automatic} have shown superior performance on this task.  

\noindent
Many methods~\cite{deshpande2015learning, larsson2016learning, zhang2016colorful, iizuka2016let} colorize images only with a grayscale. As there can be multiple plausible colorized images,~\cite{deshpande2017learning, messaoud2018structural, wu2021towards, kim2022bigcolor} explores generating diverse colorization. Some of these methods use generative priors for diverse colorization. These methods \cite{vitoria2020chromagan, su2020instance, zhao2020pixelated} use semantic information for better plausible colorization.

\noindent
\textbf{Video Colorization.} Compared to image colorization, video colorization is more challenging as it has to color an entire sequence while maintaining temporal consistency along with spatial consistency.~\cite{lei2019fully} introduces an automatic approach for video colorization with self-regularization and diversity without using any label data.~\cite{zhang2019deep} presents an exemplar-based method that is temporally consistent and remains similar to the reference image. They use a recurrent framework using semantic correspondence and color propagation from the previous step.

\noindent
\textbf{3D Representations.} NeRF~\cite{mildenhall2021nerf} has become a popular choice of 3D representation for novel-view synthesis tasks. Representations like InstantNGP~\cite{mueller2022instant}, Plenoxels~\cite{fridovich2022plenoxels}, DVGO~\cite{SunSC22}, Mip-NeRF360~\cite{barron2022mip}, Zip-NeRF~\cite{barronzip} have enhanced the original NeRF~\cite{mildenhall2021nerf} by reducing aliasing, training and rendering time. Recently, Gaussian-Splatting~\cite{kerbl3Dgaussians}, which uses rasterization of splats instead of volumetric rendering, was introduced, accelerating training and achieving real-time rendering. Further, these representations have become popular for solving other tasks such as dynamic scenes~\cite{liu2023robust, yang2024deformable, park2021nerfies}, hierarchical scenes~\cite{dhiman2023strata}, text-to-3D generation~\cite{wang2024prolificdreamer, tang2023dreamgaussian}, and large-scale scenes~\cite{rematas2022urban}. 

\noindent
\textbf{Knowledge Distillation.} Hinton~\etal~\cite{hinton2015distilling} distilled the soft targets generated by a larger network to a smaller network. Some common approaches include distillation based on the activations of hidden layers in the network~\cite{heo2019knowledge}, distillation based on the intermediate representations generated by the network~\cite{aguilar2020knowledge}, and distillation using an adversarial loss function to match the distributions of activations and intermediate representations of the two networks~\cite{wang2018kdgan}.

\section{Method}

\subsection{Preliminaries}
\label{subsec:prelims}
\noindent
\textbf{NeRF.} NeRF~\cite{mildenhall2021nerf} represents the implicit 3D geometry of a scene by learning a continuous function $f$ whose input is 3D location $x$ and a viewing direction $d$ and outputs are color $c$ and volume density $\sigma$, which is parameterized by a multi-layer perceptron (MLP) network. During rendering, a ray $r$ is cast from the camera center along the viewing direction $d$ and is sampled at different intervals. Then, NeRF estimates the color of a pixel by weighted-averaging of the colors of sampled 3D points using volumetric rendering~\cite{mildenhall2021nerf}. The MLP is learned by optimizing the squared error between the rendered pixels $f(r)$ and the ground truth pixels $I(p)$ from multiple input views:
\begin{equation}
    \label{eq:photo_loss}
    L_{photo} = {|| I(p) - f(r)||}_2^2
\end{equation} 

\noindent
\textbf{\textbf{Hybrid Representations.}} Recently, hybrid representations like InstantNGP~\cite{mueller2022instant}, Plenoxels~\cite{fridovich2022plenoxels}, and DVGO~\cite{SunSC22} have become popular as they use grid-based representation, which is much faster than the traditional NeRF representations. We develop upon Plenoxels~\cite{fridovich2022plenoxels}, which represents a 3D scene with sparse voxel grids, and learn spherical harmonics and density for each voxel grid. Spherical harmonics are estimated for each of the color channels. For any arbitrary 3D location, density and spherical harmonics are trilinearly interpolated from the nearby voxels. Plenoxels also use the photometric loss described in NeRF~\cite{mildenhall2021nerf} (Eq.~\ref{eq:photo_loss}). Additionally, they also use total variation (TV) regularization on the voxel grid. The final loss is as follows:
\begin{equation}
    \label{eq:plenoxel_loss}
    L_{rendering} = L_{photo} + \lambda_{TV} L_{TV}
\end{equation}
\vspace{-2em}
\subsection{Overview}
\noindent
Given a set of multi-view grayscale 
images of a scene $X = \{X_1, ..., X_n\}$ and corresponding camera poses $P = \{P_1, ..., P_n\}$, we learn a radiance field network $f_{\theta}$ which predicts density $\sigma$ and color $c$ along a camera ray $r$. To achieve this, we propose a two-stage learning framework. Even though the input to the radiance field network is multi-view grayscale images, we can still learn the underlying geometry and luminance of the scene. This is the first stage in our pipeline: \textit{``Luma Radiance Field Stage''} which learns the geometry and luminance of the 3D scene. Next, we distill the knowledge from a pre-trained colorization network trained on natural images to the learned radiance field network in the previous stage. This is \textit{``Color Distillation Stage''} in our method. Fig.~\ref{fig:method} illustrates the overall pipeline of our method. We discuss\textit{``Luma Radiance Field Stage''} in Section~\ref{subsec:stage1} and  \textit{``Color Distillation Stage''} in Section~\ref{subsec:stage2}.

\subsection{Luma Radiance Field Stage}
\label{subsec:stage1}
\noindent
We train a neural radiance field network using Plenoxels~\cite{fridovich2022plenoxels} $f_{\theta}$ to learn the implicit 3D function of the scene. As our method does not have access to the color image, we take photometric loss w.r.t to the ground-truth grayscale image following Eq.~\ref{eq:photo_loss}. 
 We observe that it has no issues in learning the grayscale images, both qualitatively and quantitatively. ( Appendix~\ref{sec:app_grey_novel_views} in the supplementary material)

\subsection{Color Distillation Stage}
\label{subsec:stage2}
\noindent
From the previous stage, we have a trained radiance field $f_{\theta}$, which has learned the implicit 3D function of the scene but generates grayscale novel views. Directly updating color information in the learned implicit 3D function is not possible. To update the implicit 3D representation, we must compute the loss on the rendered view. 
%MLP parametrizes the 3D scene as a set of weights and optimizes this MLP using the rendering loss. Hence, it is impossible to directly update the color location of a 3D point in these representations. Directly updating the color of a 3D location in hybrid representations is also not feasible because the color is parametrized using spherical-harmonic coefficients for view-dependent effects. 
Therefore, the optimal strategy for colorizing a radiance field network is to distill knowledge from pre-trained colorization networks trained on a large dataset of natural images. 

\begin{algorithm}[!t]
 \caption{Color Distillation With Multi-Scale Regularization (Appendix~\ref{sec:app_multiscale_notations} for notations)}
 \label{algo:multi-scale}
		 \textbf{Input}: Trained NeRF model $f_{\theta}$ on multi-view grayscale images, colorization teacher network $\mathcal{T}$ \\
		 \textbf{Output}: Colorized NeRF model
		 \begin{algorithmic}[1]
			 \Function{Loop}{for each image i=1,2.....N do}\\ \indent
			     \indent $\mathcal{L}_{i} \leftarrow \phi$ \\ \indent
                 \indent $I^{C}_{i} \leftarrow  \mathcal{T}(X_{i})$. \\ \indent
			     \indent $\mathcal{P}_a \leftarrow \phi$ \\ \indent
			     \indent $\mathcal{P}_b \leftarrow \phi$ 
			   \Function{Loop}{for each scale s=K,...,1,0 do}\\ \indent
			       \indent $^{s}I^{C}_{i} \leftarrow downsample(I^{C}_{i}, 2^{s})$. \\ \indent
			       \indent $^{s}I^{R}_{i} \leftarrow f_{\theta}(P_i, s)$ \\ \indent
			       \indent $\mathcal{L}_{i} \leftarrow  \mathcal{L}_{i} + \mathcal{L}_{distill}(^{s}I^{C}_{i}, ^{s}I^{R}_{i})$ .
			       \Function{If}{s != $K$}\\ \indent 
			       \indent \indent $\mathcal{L}_{i} \leftarrow \mathcal{L}_{i} +  ||\mathcal{P}_a - ^{s}a^{R}_{i}|| + ||\mathcal{P}_b - ^{s}b^{R}_{i}||$\indent\EndFunction \\ \indent
			         \indent $\mathcal{P}_a \leftarrow upsample( ^{s}a^{R}_{i}, 2)$ \\ \indent
			        \indent $\mathcal{P}_b \leftarrow upsample( ^{s}b^{R}_{i}, 2)$ \\ \indent
			   \indent Update $f_{\theta} $
			     \EndFunction
			  \EndFunction 
			 \end{algorithmic}
\end{algorithm}

\noindent
We propose a color distillation strategy that transfers color details to a 3D scene parameterized by $f_{\theta}$ from any image colorization network $\mathcal{T}$ trained on natural images. More precisely, given a set of multi-view grayscale images of a scene $X = \{X_1, ..., X_n\}$, we pass them through the colorization network $\mathcal{T}$ to obtain a set of colorized images $I^C = \{I^{C}_{1}, I^{C}_{2}, ... , I^{C}_{n}\}$. Corresponding to the camera poses of these images, we obtain rendered images $I^R = \{I^{R}_{1}, I^{R}_{2}, ... , I^{R}_{n}\}$ from $f_{\theta}$ trained in the previous stage on $X$.  We convert both $I^{C}_{i}$ and $I^{R}_{i}$ to \textit{Lab} color space and distill knowledge from the color network $\mathcal{T}$. Then, our distillation loss, $\mathcal{L}_{D}$, can be written as :
\begin{multline}
    \label{eq:distillation}
    \mathcal{L}_{D}(I^{C}_{i}, I^{R}_{i}) = ||L^{C}_{i} - L^{R}_{i}||^2 + 
    ||a^{C}_{i} - a^{R}_{i}|| + ||b^{C}_{i} - b^{R}_{i}|| 
 \end{multline}

\noindent
To summarize, we minimize MSE loss between the luma channel and use L1 loss for \textit{a} and \textit{b} channels. MSE loss preserves the content of the original grayscale images and L1 loss on the chroma channels distills the color from the colorization network. We briefly summarize this color distillation in Appendix~\ref{sec:app_overview_color_distillation} in the supplementary material.

\begin{figure*}[!t]
    \centering
    \includegraphics[width=\linewidth]{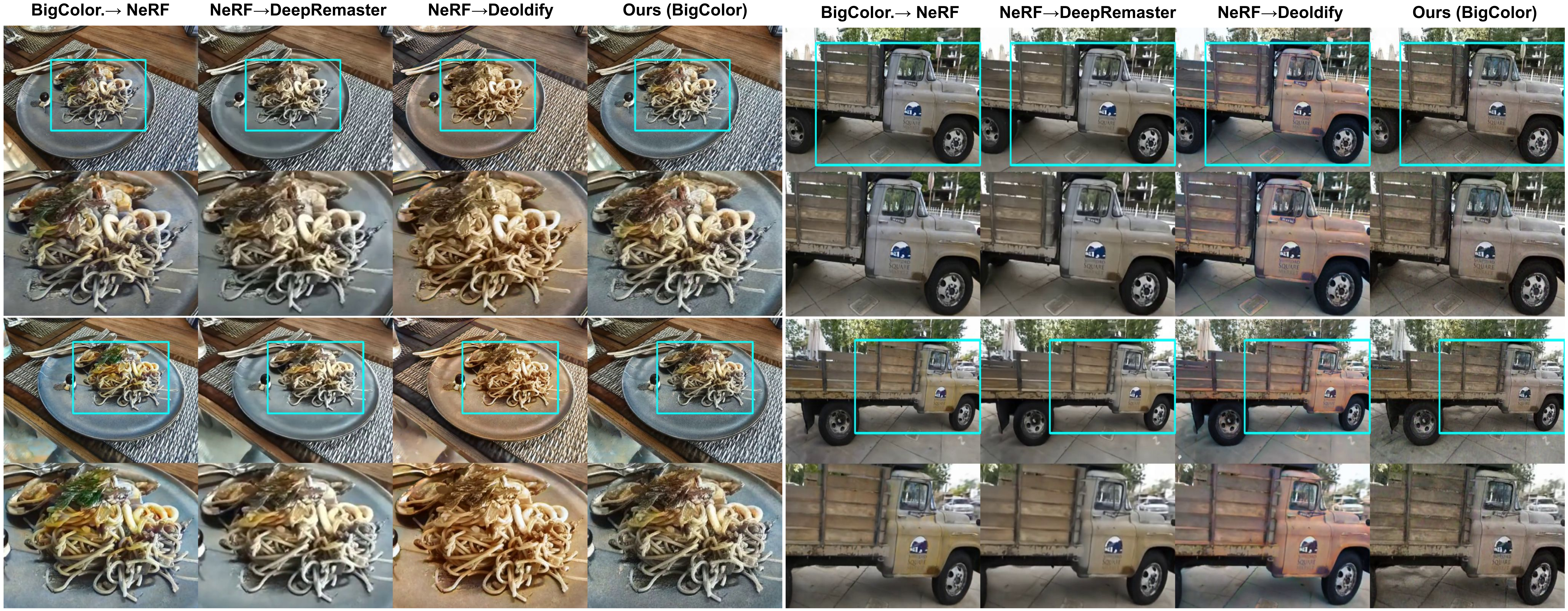}
    \caption{\textbf{Qualitative results of our method on baselines for ``Pasta'' and ``Truck'' scene.} We display two novel views rendered from different viewpoints, with rows 1 and 3 at the original resolution and rows 2 and 4 zoomed in on the highlighted regions. Even the video-based baselines (columns 2 and 3) exhibit inconsistencies. Note the color change in highlighted regions in ``Truck'' scene.}
    \label{fig:qual_results_video}
    \vspace{-1em}
\end{figure*}

\noindent
\textbf{Multi-scale regularization.} Sometimes, colorized views appear to desaturated or washed out. To mitigate this, we introduce multi-scale regularization in the colorized views. In multi-scale regularization, we analyze an image at different scales by constructing image pyramids that correspond to different scales of an image. The lowest level of the pyramid contains the image structure and dominant color, while the finer level, as the name indicates, contains finer features. We create an image pyramid by progressively sub-sampling an image. Then, we start color distillation at the coarsest scale, as discussed in the previous section. For subsequent scales, we regularize the predicted chroma channels with the prediction from the previous scale. We provide details about this regularization in Algorithm~\ref{algo:multi-scale}.
$\mathcal{P}_a$ and $\mathcal{P}_b$ are placeholders to keep the interpolated predicted chroma channels from the previous scale. We use bilinear interpolation to upsample the chroma channels. Distilling color from coarsest-to-finest levels ensures prominent colors are learned during optimization, which mitigates the desaturation in the colorized views. We provide ablation in Appendix~\ref{sec:supp_ms_regularization} in the supplementary material. 

\noindent
\textbf{Implementation Details} We use Plenoxel as our radiance field network representation (Section~\ref{subsec:prelims}). We use the loss described in Eq.~\ref{eq:plenoxel_loss} for the datasets used in our experiments.  During the Color Distillation stage, we estimate the loss in \textit{Lab} color space as described in Eq.~\ref{eq:distillation}. %We use the deferred back propagation technique proposed by ARF~\cite{zhang_arf} to backpropagate the loss. 
We train Color Distillation stage only for $10$ epochs. 
\section{Experiments}
\noindent
This section presents qualitative (Section~\ref{subsec:qual_results}) and quantitative (Section~\ref{subsec:quant_results}) experiments to evaluate our method. Our method's effectiveness is demonstrated with two image colorization teacher networks~\cite{zhang2017real} and~\cite{kim2022bigcolor}. 
%To summarize, our method takes a set of grayscale posed images of a given scene and learns to generate colorized novel views.  
We compare our approach with two trivial baselines: 1.) colorize input multi-view grayscale images and then train a radiance field network, and 2.) colorize the generated novel-view grayscale image sequence using a video colorization method. To quantitatively evaluate, we use a cross-view consistency metric using a state-of-the-art optical flow network  RAFT~\cite{teed2020raft} discussed in~\cite{nguyen2022snerf, huang2022stylizednerf}. Further, we also compare our method with a concurrent work and show that the stylization methods are unsuitable for the colorization task in 3D.

\noindent
In addition, we conducted a user study to qualitatively evaluate the colorization results. We also present ablation for multi-scale regularization in Appendix~\ref{sec:supp_ms_regularization} in the supplementary material. Finally, we show results on two real-world downstream applications - colorization of radiance field networks trained on 1.) Infra-Red (IR) and 2.)In-the-wild grayscale images. Our experiments show that our approach outperforms the baseline methods, producing colorized novel views while maintaining 3D consistency. 
%Our strategy ensures 3D consistency in NeRF colorization by leveraging advancements in image colorization networks. 
We encourage readers to watch the supplementary video to better assess our work. 

\noindent
\textbf{Datasets.} We conduct experiments on two types of real-scenes: \RNum{1}) forward-facing real scenes  LLFF~\cite{mildenhall2019llff} and Shiny~\cite{shiny} dataset; and \RNum{2}) $360^{\circ}$ unbounded real-scenes Tanks \& Temples (TnT)~\cite{knapitsch2017tanks} dataset.   LLFF~\cite{mildenhall2019llff} dataset provides $24$ scenes captured using a handheld cellphone, and each scene has $20-30$ images. 
%The camera poses are extracted through COLMAP~\cite{schoenberger2016mvs}. 
Shiny~\cite{shiny} has $8$ scenes with multi-view images. Tanks \& Temples (TnT)~\cite{knapitsch2017tanks} also has $8$ scenes that are captured in realistic settings with an industry-quality laser scanner for capturing the ground truth. These datasets have a variety in terms of objects, lighting, and scenarios. For experimentation purposes, we convert the images in the dataset to grayscale. 
%We use the resolution size per the recommended configuration files in Plenoxel~\cite{fridovich2022plenoxels}.

\noindent
\textbf{Baselines.}We compare with the following baselines:
\vspace{-0.125em}
\begin{enumerate}
    \item \textbf{Image Colorization $\rightarrow$ Novel View Synthesis.} : Train Plenoxels~\cite{fridovich2022plenoxels} on colorized images using image colorization method~\cite{zhang2016colorful, kim2022bigcolor}.
    \item \textbf{Novel View Synthesis $\rightarrow$ Video Colorization}: Train Plenoxels~\cite{fridovich2022plenoxels} on grayscale images and obtain colorized novel-views by applying state-video colorization methods~\cite{iizuka2019deepremaster, salmona2022deoldify} to the rendered images.
\end{enumerate}

\begin{figure}[!t]
    \centering
    % \includegraphics[width=\linewidth]{Figures/Style_Comparison.pdf}
    
    % \caption{Results from (a) ARF~\cite{zhang2022arf} and (c) Our method. (b) Zoomed-in region of (a) and (d) Zoomed-in region of (c). Check the perceptual artifacts from results in color-ARF.}
    \includegraphics[width=\linewidth]{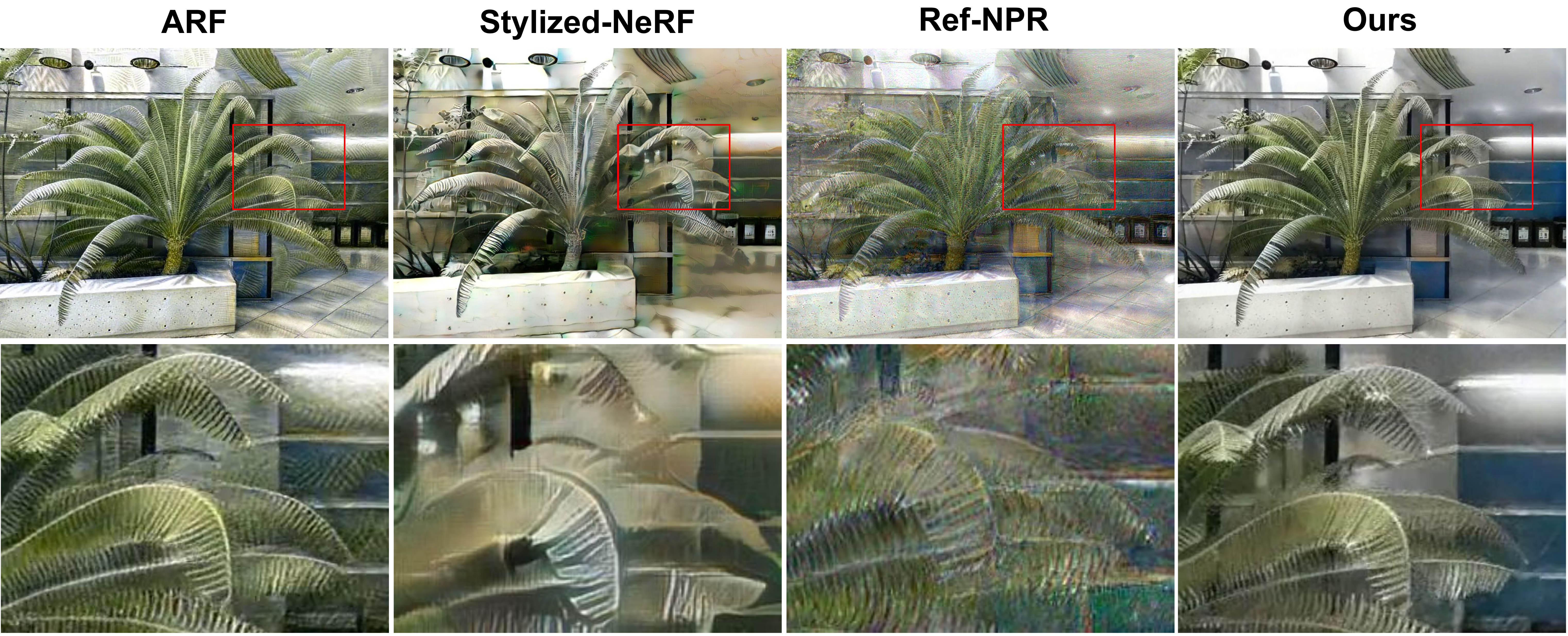}
    
    \caption{(left-to-right) Results from ARF~\cite{zhang2022arf}, Stylized-NeRF~\cite{huang2022stylizednerf}, Ref-NPR~\cite{zhang2023ref} and Our method. (Bottom Row) Zoomed-in region of the highlighted region. Check the artifacts from results in stylization works}

    \label{fig:comparison_style}
    \vspace{-1em}
\end{figure}

%All baselines use the same radiance field representation: Plenoxel. 

\noindent
For baseline 1, we use~\cite{zhang2017real} and~\cite{kim2022bigcolor} for colorizing the input views, thus creating two versions for this baseline. Similarly, for baseline 2, we create two versions using DeepRemaster~\cite{iizuka2019deepremaster} and DeOldify~\cite{salmona2022deoldify}. 
%We did not use image colorization techniques on the rendered grayscale views because they do not consider temporal and multi-view consistency. Hence, results will not be multi-view consistent. Similarly, we did not apply video-colorization techniques to the multi-view grayscale images because different input views could lead to different sequences for the video-colorization network. 
Further, we compare our results with stylization works and a contemporary work Color-NeRF~\cite{cheng2024colorizing}. 

\subsection{Qualitative Results}
\label{subsec:qual_results}
\noindent
\textbf{Image Colorization $\rightarrow$ Novel View Synthesis.} We compare our method with both versions of this baseline in Fig.~\ref{fig:qual_results}.  We generate novel views from two different viewpoints to facilitate a better comparison of the 3D consistency. The baselines exhibit significant color variation in the ``Cake'' scene, while our method produces results without color variation. Similarly, for ``Leaves'' and ``Pasta'' scenes, color variations can be observed in the highlighted region. We also observe similar 3D consistency in the TnT~\cite{knapitsch2017tanks} dataset, as shown in Fig.~\ref{fig:qual_results} in the bottom two sets. Our method visually demonstrates better 3D consistency in the generated novel views.

\noindent
\textbf{Novel View Synthesis $\rightarrow$ Video Colorization.} We compare with the video-colorization-based baseline in Fig.~\ref{fig:qual_results_video} for the ``Pasta'' scene from LLFF~\cite{mildenhall2019llff} dataset and the ``Truck'' scene from TnT~\cite{knapitsch2017tanks} dataset. The video-based baseline shows better consistency than the image-based baseline but still produces inconsistent colorization. Our method preserves consistency due to explicit modeling in 3D. We can observe a color change in the plate and truck body from Deoldify~\cite{salmona2022deoldify} baseline version. Our method preserves color consistency on the truck body and plate across two views. 

\noindent
\textbf{Comparison with NeRF-Stylization methods.} We also compare our method with NeRF-stylization methods ARF~\cite{zhang2022arf}, Stylized-NeRF~\cite{huang2022stylizednerf} and Ref-NPR~\cite{zhang2023ref} by giving a color image as a style image. We observe artifacts in results from these stylization methods in Fig.~\ref{fig:comparison_style}. Stylization involves transferring the overall style of one image to another image or video, focusing on overall texture differences using loss functions like LPIPS. In contrast, colorization emphasizes achieving plausible colors by accurately representing local color values. Therefore, stylization techniques are unsuitable for the colorization of radiance fields. 

%The stylization task involves transferring the overall style of one image to another image or video. For instance, a prominent loss function used in stylization work is LPIPS, which primarily penalizes differences in overall texture rather than local color values. On the other hand, the colorization task prioritizes achieving plausible colors, focusing on accurately representing local color values. Hence, stylization works cannot be utilized for the colorization task for radiance fields. 

%\textbf{Novel View Synthesis.} We show additional qualitative and quantitative results in Appendix C.2 of the supplementary material. Our method maintains 3D consistency across all views despite challenging lighting conditions. These results show generalization of our method across different scenes.

\begin{figure}
    \centering
    \includegraphics[width=\linewidth]{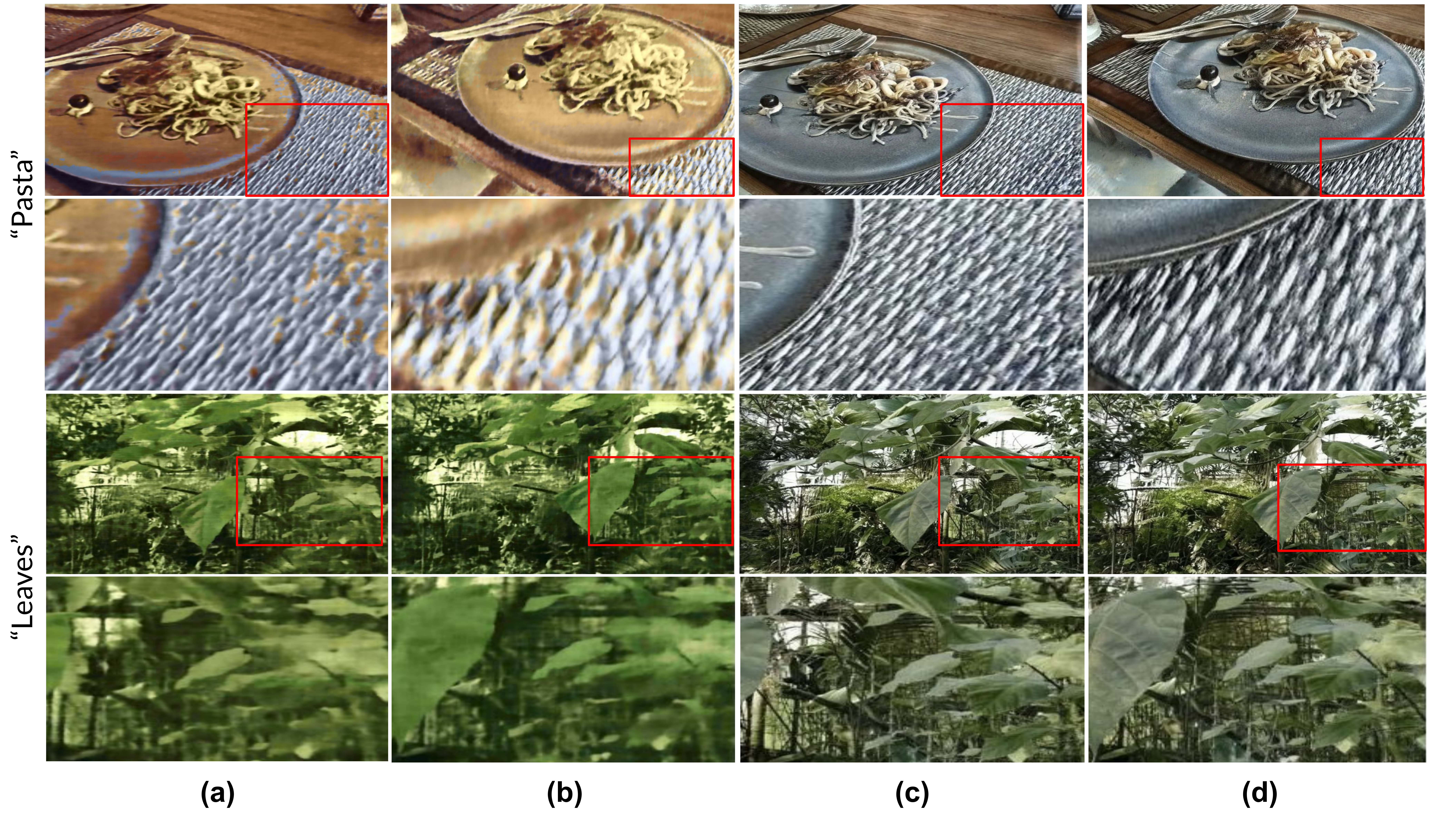}
    \caption{(a) \& (b) Novel-views from Color-NeRF~\cite{cheng2024colorizing} and (c) \& (d) Novel-views from our method. Bottom row of each scene illustrates zoomed-in regions. Notice the inconsistency in Color-NeRF.}
    \label{fig:qual_aaai_cmp}
    \vspace{-4mm}
\end{figure}

\begin{figure}[!t]
    \centering
    \includegraphics[width=\linewidth]{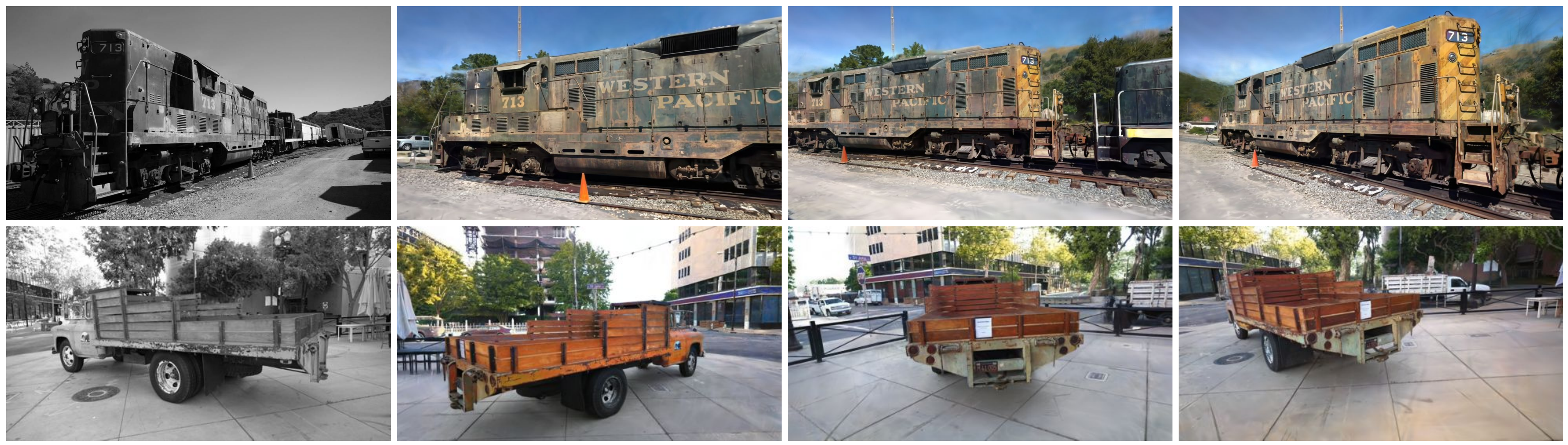}
 \caption{ (First column) Grayscale novel-view. Colorized novel-views from our method with Gaussian-Splatting~\cite{kerbl3Dgaussians} backbone for ``Train''(Top) and ``Truck''(Bottom)  scenes. These results demonstrate that our method maintains multi-view consistency and extends seamlessly to rasterization-based 3D representation.}
\vspace{-10pt}
\label{fig:gs_static}
\end{figure}

\noindent
\textbf{Comaprison with Color-NeRF~\cite{cheng2024colorizing}} We compared our method against a concurrent work Color-NeRF. Fig.~\ref{fig:qual_aaai_cmp} shows qualitative results for a forward-facing scene from LLFF. We observe that the novel-views from Color-NeRF are not consistent. Notice the color change on the plate. In comparison, results from our method are consistent. Further, it takes nearly 10 hours end-to-end on RTX A6000 for a forward-facing scene for Color-NeRF. Compared to this, our method takes only 1 hour with Plenoxel backbone and 30 minutes with 3DGS backbone. We produce more comparison results in Appendix~\ref{sec:supp_comparison_colornerf}.

\noindent
\textbf{Results with Gaussian-Splatting~\cite{kerbl3Dgaussians} backbone} Our method can be extended to Gaussian-Splatting representation, which uses rasterization of Gaussian splats for rendering. We provide details for training with Gaussian-Splatting as backbone in Appendix~\ref{sec:supp_training_3dgs} of the supplementary material. Fig.~\ref{fig:gs_static} provides qualitative results on two scenes: ``Train'' and ``Truck'' from TnT~\cite{knapitsch2017tanks} dataset. Similar to the backbone with Plenoxels, we observe that the colorized novel-views are multi-view consistent. 

% \input{figures_latex/dynamic_results}
% \textbf{Dynamic Scenes} We also test the efficacy of the proposed method on dynamic scenes i.e. scenes in which local motion is happening. We use Deformable-3D-Gaussians~\cite{yang2023deformable3dgs} as the backbone representation. We provide more details on this training in the supplementary material. We observe that our method is able to produe multi-view consistent results even for dynamic scenes. Fig.~\ref{fig:gs_dynamic} shows a sequence of colorized novel-views for ``Chicken'' scene from Nerfies~\cite{park2021nerfies} dataset. 

\begin{table}[!t]
\caption{Quantitative results for cross-view short-term and  long-term consistency on LLFF dataset.}
\vspace{-2mm}
\label{tab:quantitative}
\begin{adjustbox}{width=\linewidth}
\begin{tabular}{@{}c|cccc|cccc@{}}
\toprule
\multicolumn{1}{c|}{} & \multicolumn{4}{c|}{\textbf{Short-Term Consistency $\downarrow$}} & \multicolumn{4}{c}{\textbf{Long-Term Consistency $\downarrow$}} \\ \midrule
                      &  \textbf{Cake} & \textbf{Pasta} & \textbf{ Buddha} & \textbf{Leaves}  & \textbf{Cake} & \textbf{Pasta} & \textbf{Buddha} & \textbf{Leaves}   \\ \midrule
      \textbf{BigColor $\rightarrow$ NeRF}        & 0.037 & 0.030 & 0.022 & 0.015 & 0.060 & 0.039 & 0.033 & 0.024     \\
      \textbf{NeRF $\rightarrow$ DeepRemaster} & 0.018 & 0.015 & 0.015 & 0.015 & 0.032 & 0.023 & 0.023 & 0.021 \\
\textbf{NeRF $\rightarrow$ DeOldify} & 0.023 & 0.034 & 0.017 & 0.032 & 0.033 & 0.049 & 0.022 & 0.040 \\
\textbf{Ours(Colorful Image Colorization)}       & \textbf{0.009} & \textbf{0.009} & \textbf{0.008} & 0.009  & \textbf{0.013} & \textbf{0.017} & \textbf{0.012} & 0.015 \\
\textbf{Ours(BigColor)}       & 0.019 & 0.015 & 0.015 & \textbf{0.008} & 0.033 & 0.025 & 0.023 & \textbf{0.013}
\end{tabular}
\end{adjustbox}
\vspace{-4mm}
\end{table}

\subsection{Quantitative Results.}
\label{subsec:quant_results}

\noindent
\textbf{Measurement of 3D consistency.} To evaluate the 3D consistency across generated novel views, we adopt a strategy proposed by ~\cite{lai2018learning}, which is also used by various NeRF-based stylization methods~\cite{nguyen2022snerf,huang2022stylizednerf}. First, we render novel views from the colorized radiance field. We need optical flow and occlusion masks between two views to compute the metric. The occlusion mask, denoted as $M$, represents occluded regions, out of bounds, or with motion gradients. We use RAFT~\cite{teed2020raft} to predict the optical flow between two views. Then, we warp a rendered view $I_i$ to obtain a warped view $\hat{I_{i+\Delta}}$; where $\Delta$ is the frame-index offset. Consistency error is defined as: 
\vspace{-5pt}
\begin{equation}
E_{\text {consistency }}\left(I_{i+\Delta}, \hat{I}_{i+\Delta}\right)=\frac{1}{\left|M\right|}\left\|{I}_{i+\Delta}-\hat{I}_{i+\Delta}\right\|^2    
\end{equation}

Similar to~\cite{nguyen2022snerf,huang2022stylizednerf} we show this metric on short-range and long-range pairs. For color consistency, we measure error only in the chroma channels.
 
\begin{table}[!t]
\vspace{-2mm}
%\begin{table}[!t]
\caption{Quantitative results for cross-view long-term consistency on Tanks \& Temples dataset.}
\label{tab:quantitative_long_term}
\begin{adjustbox}{width=\linewidth}
\begin{tabular}{@{}c|cccc@{}}
\toprule
\textbf{Long-Term Consistency $\downarrow$} & \textbf{Horse} & \textbf{M60} & \textbf{Train} & \textbf{Truck} \\ \midrule
\textbf{Colorful Image Colorization $\rightarrow$ NeRF}         & 0.018 & 0.017 & 0.028 & 0.035 \\
\textbf{BigColor$\rightarrow$ NeRF}        & 0.022 & 0.027 & 0.034 & 0.038 \\ \midrule
\textbf{NeRF $\rightarrow$ DeepRemaster} & \textbf{0.017} & 0.022 & \textbf{0.024} & 0.032 \\
\textbf{NeRF $\rightarrow$ DeOldify} & 0.032 & 0.031 & 0.025 & 0.031 \\ \midrule
\textbf{Ours(Colorful Image Colorization)}       & 0.018 & \textbf{0.015} & 0.026 & \textbf{0.020} \\
\textbf{Ours(BigColor)}       & 0.020 & 0.021 & 0.031 & 0.028 \\ \bottomrule
\end{tabular}
\end{adjustbox}
%\end{table}
\vspace{-4mm}
\end{table}

\begin{figure}[!t]
  \centering
\begin{minipage}[c]{0.48\linewidth}
    \centering
    \includegraphics[width=\linewidth]{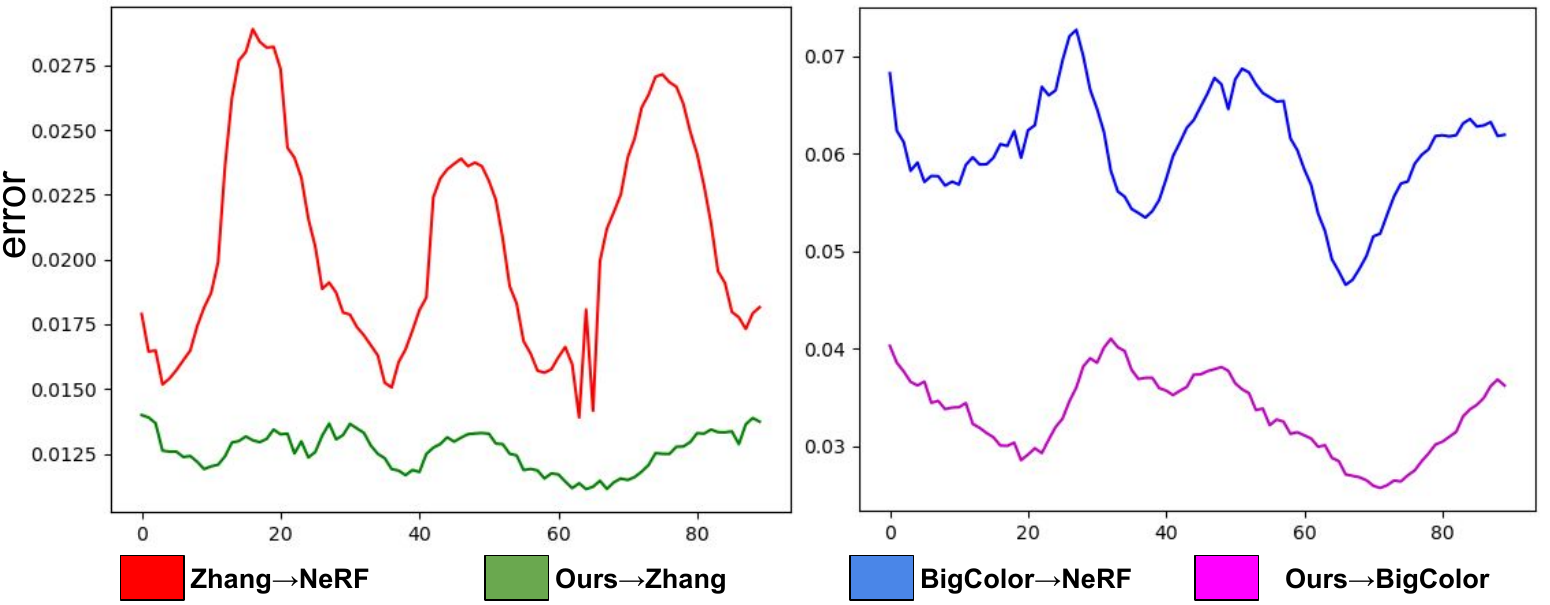}
    
    \caption{Metrics distribution for (Left)~ \cite{zhang2016colorful} and (Right) BigColor~\cite{kim2022bigcolor} for ``cake'' scene. We observe that variation from our method has less variance compared to both versions of the image-colorization-based baseline.}
    %\vspace{-20pt}
    \label{fig:metrics_plot}
\end{minipage}  \hfill
\begin{minipage}[c]{0.48\linewidth}
\centering
    \includegraphics[width=\linewidth]{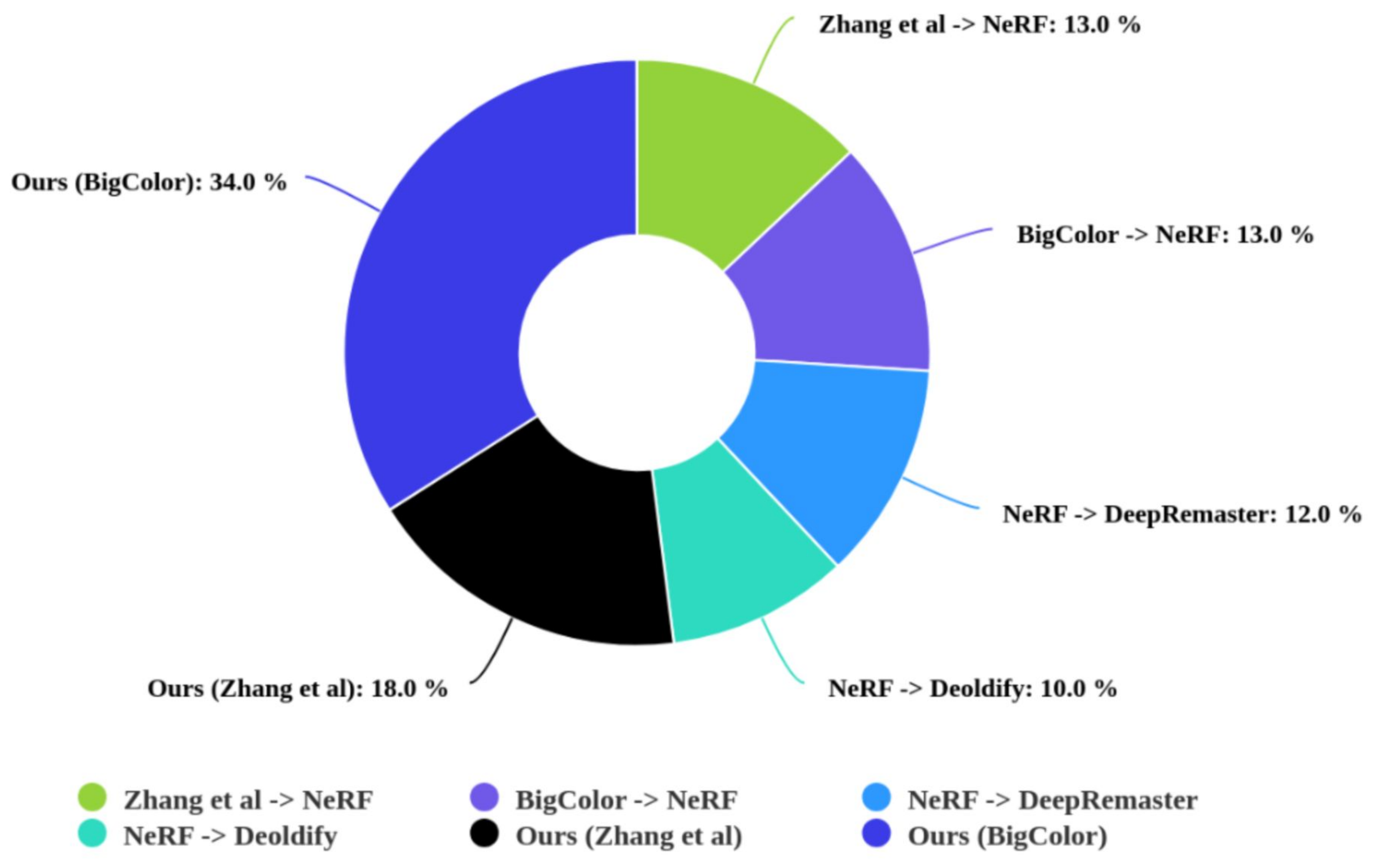}
    \caption{User Study. Our result maintains view consistency after colorization and performs better than the baselines.}
    %\vspace{-1.25em}
    \label{fig:userstudy}
\end{minipage}
\vspace{-16pt}
\end{figure}

\noindent
Table~\ref{tab:quantitative} and~\ref{tab:quantitative_long_term} show short-term and long-term consistency metrics for LLFF and TnT datasets, respectively. 
%The generated novel views in the rendering trajectory have a temporal difference of 10 in short-term consistency and 30 in long-term consistency. 
We observe that our qualitative findings align with these quantitative results. For our method with ~\cite{zhang2016colorful} and BigColor, we observe that short-term and long-term consistency improves for different scenes. For ``Pasta'' scene in Tab.~\ref{tab:quantitative} we see $70\%$ and $56.41\%$ reduction when compared with image-based baseline. We observe significant improvement in metrics when compared with different baselines. Further, our approach generates better cross-view consistent novel-views, regardless of the pre-trained colorization teacher. Additionally, our method produces more consistent novel views than video-based baselines. 

%BigColor~\cite{kim2022bigcolor} due to its more colorful views leads to significant variations across multiple perspectives. Nonetheless, 
%Further, we observe that our approach generates better cross-view consistent novel-views, regardless of the pre-trained colorization teacher. Additionally, our method produces more consistent novel views than video-based baselines. 
%For the ``Train'' scene in Table~\ref{tab:quantitative_long_term}, even though video-based baselines perform better qualitatively our method yields consistent novel-views. 
\noindent
Fig.~\ref{fig:metrics_plot} shows the distribution of metrics for the entire rendering sequence for both pre-trained models. The error curve from our method is consistently lower and smoother than the baselines, validating our claim of consistency in novel views obtained from our distillation method.   

\noindent
\textbf{User Study.} We provide users with $12$ colorized sequences from LLFF~\cite{mildenhall2019llff}, Shiny~\cite{shiny}, Shiny Extended~\cite{shiny} and Tanks \& Temples (TnT)~\cite{knapitsch2017tanks} to compare our method with baseline techniques. Users were asked to select the scene with the best view consistency, vivid color, and no color spill into neighboring regions. We invited $30$ participants and asked them to select the best video satisfying these criteria. Fig.~\ref{fig:userstudy} shows that our method was preferred $52\%$ of the time.% indicating the 3D consistency in our method. 

\begin{figure}[!t]
    \centering
    \includegraphics[width=\linewidth]{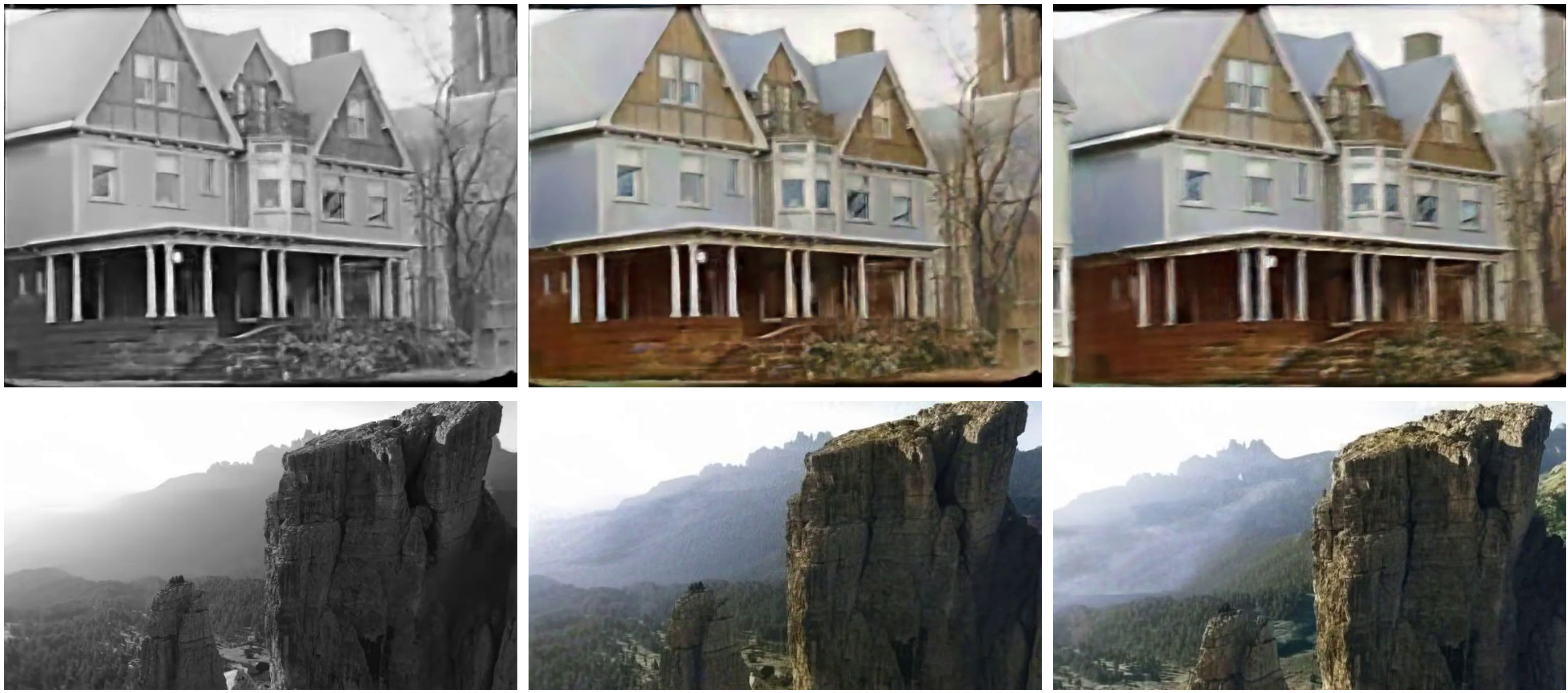}
    
    \caption{\textbf{Results on In-the-wild grayscale-sequences.} First column represents the input grayscale scene. Columns 2-3 illustrate the colorized novel-view sequence from our method. (Top Row) ``Cleveland in 1920s - House''. (Bottom Row) `` Mountain - Cinematic Video''. Our method generates consistent colorized views.}
    \vspace{-4mm}
    \label{fig:legacy_app}
\end{figure}

\begin{figure}[!t]
    \centering
    \includegraphics[width=\linewidth]{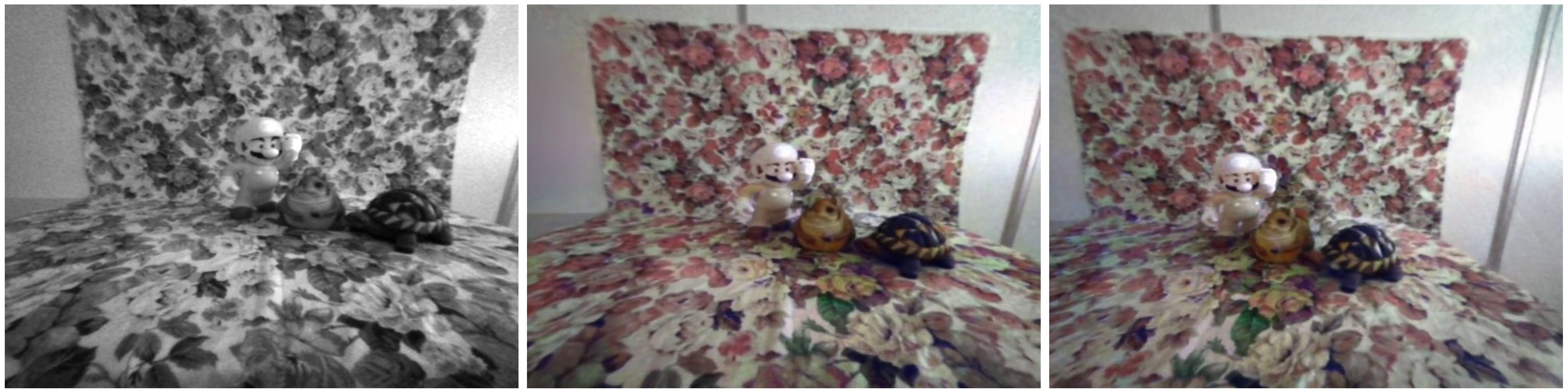}
 \caption{(Column 1) Input multi-view IR Sequence. (Columns 2 and 3) Colorized multi-views from Our method. Our approach yields consistent novel-views for a different input modality.}  
\label{fig:ir_app}
\vspace{-4mm}
\end{figure}

\section{Applications}
\noindent
\textbf{Multi-View IR images.} Our method is highly significant for modalities that do not capture color information. One such popular modality is IR images. For this experiment, we obtain data from~\cite{poggi2022cross}. This dataset is generated from a custom rig consisting of IR and multi-spectral (MS) sensor and RGB camera. This dataset contains $16$ scenes and $30$ views per modality. We show novel views in Fig.~\ref{fig:ir_app}. We observe that a teacher trained on natural images works well for colorizing the scene. We also discuss the benefits of colorization for the object-detection task in Appendix~\ref{sec:suppl_downstream} in the supplementary material.

\noindent
\textbf{In-the-wild grayscale images.} We demonstrate our approach's capability to colorize real-world old videos. We extract an image sequence from an old video: ``Cleveland in 1920s'' and pass them through COLMAP~\cite{schonberger2016structure} to extract camera poses. Then, we use our framework to generate the color novel views from this grayscale legacy content input. Similarly, we generate novel views for ``Mountain'' sequence. We can observe in Fig.~\ref{fig:legacy_app}  that our method can get 3D consistent novel views for such sequences. This is useful in the restoration of the old legacy content.

\begin{figure*}[!t]
    \centering
    \includegraphics[width=0.955\linewidth]{Figures/Qualitative_Results.pdf}
    \caption{\textbf{Qualitative results of our method with image-colorization baselines.} We display two rows of each scene, each rendered from a different viewpoint. The first four columns depict the original resolution results, while the last four columns show zoomed-in regions of the highlighted areas in the first four columns. The image-based baselines have color inconsistencies in their results, whereas our distillation strategy (columns 3, 4, 7, 8) maintains color consistency across different views.}
    \label{fig:qual_results}
    \vspace{-1em}
\end{figure*}

\section{Conclusion}
\vspace{-2mm}

\noindent
We present~\methodname{}, a novel method for colorizing radiance field networks trained on multi-view grayscale images. We use a distillation framework that leverages pre-trained colorization networks on natural images, ensuring superior 3D consistency compared to baseline methods. Multi-scale self-regularization prevents color desaturation during distillation. Our experiments demonstrate robustness to variations in teacher networks. Generated novel views from our method exhibit greater 3D consistency than baselines. Additionally, our method extends seamlessly to rasterization-based representations. A user study showed a preference for our approach. We also demonstrate applications to multi-view IR sensors and legacy image sequences.

%Add acknowledgement here
%\noindent \textbf{Acknowledgement.} Ankit Dhiman was supported by Samsung R\&D Institute India, Bangalore. R Srinath and Srinjay Sarkar were supported by Kotak IISc AI-ML Centre.

%\title{ChromaDistill: Colorizing Monochrome Radiance Fields with Knowledge Distillation}
%\author{Supplementary Material}
%\maketitle
%\maketitle

%\maketitlesupplementary

\appendix

\tableofcontents

\addtocontents{toc}{\protect\setcounter{tocdepth}{2}}

\section{Introduction}
We present additional results and other details related to our proposed method: ChromaDistill. We present training details in Appendix ~\ref{subsec:train_details}. We explain the downstream applications in Appendix ~\ref{subsec:ir_training} and ~\ref{subsec:in_the_wild}. We present additional experimental results in Appendix \ref{sec:experimental_results}.

\section{Implementation Details}
\subsection{Training Details}
\label{subsec:train_details}
We use Plenoxels~\cite{fridovich2022plenoxels} as neural radiance field representation in our experiments.  This representation uses a sparse 3D grid based representation with spherical harmonic (SH) coefficients. For the first stage, luma radiance field, we use the default Plenoxel grid recommended for the type of dataset. We use batch-size of 5000 with RMSProp as optimizer. In the first stage, we use both photometric losses and total-variation (TV) loss proposed in the plenoxels~\cite{fridovich2022plenoxels}. In the distillation stage, first we get the colorized images from the teacher network. In our experiments, we present result with two image-colorization teachers : 1.) Zhang \etal ~\cite{zhang2016colorful} and 2.) Bigcolor~\cite{kim2022bigcolor}. These colorized images are then used in the distillation stage. When distilling color, we convert the colorized image to ``Lab'' color space. 

\subsection{Infra-Red Muli-Views}
\label{subsec:ir_training}
Multi-spectral or Infra-red (IR) sensors are more sensitive to the fine details available in the scene than RGB sensors. Poggi \etal ~\cite{poggi2022cross} proposed Cross-spectral NeRF (X-NeRF) to model a scene using different spectral sensors. They built a custom rig with a high-resolution RGB camera and two low-resolution IR and MS cameras and captured 16 forward-facing scenes for their experiments. We extracted IR multi-view images and camera poses from the proposed dataset. We naively normalize the IR view between 0 and 1; thus treating it as a grayscale multi-view input sequence. We then apply our method to colorize this view. Our method is effective in colorizing views from different modalities.  

\subsection{In-the-wild GrayScale Multi-Views}
\label{subsec:in_the_wild}
Other than different multi-spectral sensors, there exist lot of in-the-wild grayscale content either in the form of legacy old videos or monochromatic cameras. We extract these multi-view image sequences and then pass these images through COLMAP~\cite{schonberger2016structure} to extract camera poses. For legacy grayscale image sequences, as there are lot of unnecessary artefacts which affects the perfomrance of COLMAP~\cite{schonberger2016structure}, we pass this sequnce through the video restoration method proposed in ~\cite{wan2022oldfilm}. We use the extracted camera-pose and grayscale multi-view image sequence as input for the propsoed method and obtain 3D consistent color-views. This downstream task has a lot of application in Augmented-reality(AR)/Virtual Reality (VR). 

\begin{algorithm}[t]

\caption{Color Distillation Algorithm}\label{algo:overview}

\textbf{Input}: Trained Nerf Model on Multi-view Grayscale images $f_{\theta}$, colorization teacher network $\mathcal{T}$ \\
\textbf{Output}: Colorized radiance field network $ \mathcal{T}$.

\begin{algorithmic}[!t]
\Function{Loop}{for each image i=1,2.....N do}\\ \indent
  \indent $\mathcal{L}_i \leftarrow \phi$ \\ \indent 
  \indent $I^{C}_{i} \leftarrow  \mathcal{T}(X_{i})$. \\ \indent 
  \indent $I^{R}_{i} \leftarrow f_{\theta}(P_i)$ \\ \indent 
  \indent $\mathcal{L}_{i} \leftarrow \mathcal{L}_{i} + \mathcal{L}_{distill}(I^{C}_{i}, I^{R}_{i})$ \indent\\ 
  \indent Update $f_{\theta} $ 
\indent \EndFunction \\
\end{algorithmic}
\end{algorithm}

\subsection{Overview of the Color Distillation Algorithm}
\label{sec:app_overview_color_distillation}
Algorithm~\ref{algo:overview} gives an overview of the color distillation algorithm, For each camera pose, we render a view from the radiance field network trained in grayscale images $f_{\theta}$. To distill the loss, we colorize the gray-scale teacher using a teacher colorization network   

\begin{figure*}
    \centering
    \includegraphics[width=\linewidth]{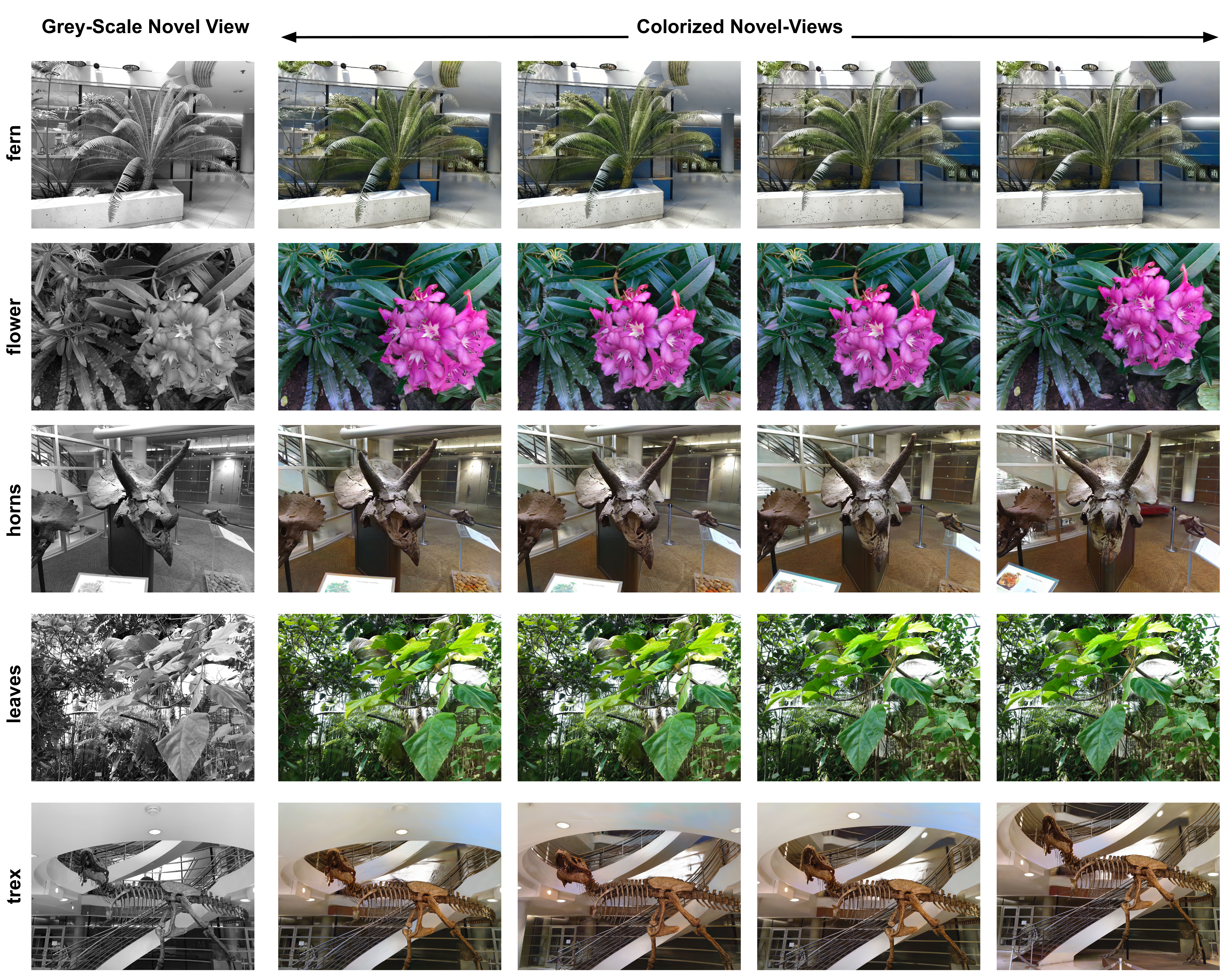}
    \caption{\textbf{Colorized Novel-views from Gausssian Splatting 3D representation,} We show results for LLFF scenes : fern, flower, horn, leaves and trex. First column is a grayscale novel-view followed by colorized novel-views using our strategy. }
    \label{fig:gs_qual}
\end{figure*}

\subsection{Notation for ``Color Distillation With Multi-Scale Regularization''}
\label{sec:app_multiscale_notations}

\begin{itemize}
    \item $f_{\theta}$ : NeRF model trained in stage 1 on multi-view grayscale images
    \item $\mathcal{L}_{i}$ : Loss for $i^{th}$ image in training-set
    \item $\mathcal{P}_a, \mathcal{P}_a$ : Placeholder to save chroma $a$ and $b$ channels from previous scale
    \item $^{s}I^{C}_{i} \leftarrow downsample(I^{C}_{i}, 2^{s})$ : Downsample the image from pre-trained colorization at original resolution by a factor $2^{s}$
    \item $^{s}I^{R}_{i} \leftarrow f_{\theta}(P_i, s)$ : Render an image with the corresponding pose at scale $s$ i.e output width and height be downscaled by a factor $2^{s}$
    \item $\mathcal{P}_a \leftarrow upsample( ^{s}a^{R}_{i}, 2)$ : upsamples the chroma $a$ and $b$ channels for next scale by a factor of 2
    \item Our method starts from the coarsest scale $K$ i.e image resolution is downscaled by a factor of $2^{K}$
\end{itemize}

\subsection{Colorization Pipeline using Gaussian Splatting~\cite{kerbl3Dgaussians}}
\label{sec:supp_training_3dgs}
Our proposed knowledge distillation method can be further applied to alternative 3D representations such as Gaussian Splatting~\cite{kerbl3Dgaussians}, which uses rasterization rather than ray-tracing for rendering. We adhere to the default hyperparameters suggested in the original study. Training is conducted only with the luma component up to $15k$ iterations, as Gaussian densification and pruning occur only up to this point. After that, we distill the ``a'' and ``b'' channels from the teacher colorization network until $30k$ iterations. We present more qualitative results in Fig.~\ref{fig:gs_qual}. Further, we use a different teacher network for colorization model DdColor~\cite{kang2023ddcolor}.

\section{Experimental Results}
\label{sec:experimental_results}

\subsection{Grayscale Novel Views}
\label{sec:app_grey_novel_views}
\begin{figure}[!t]
    \centering
    \includegraphics[width=\linewidth]{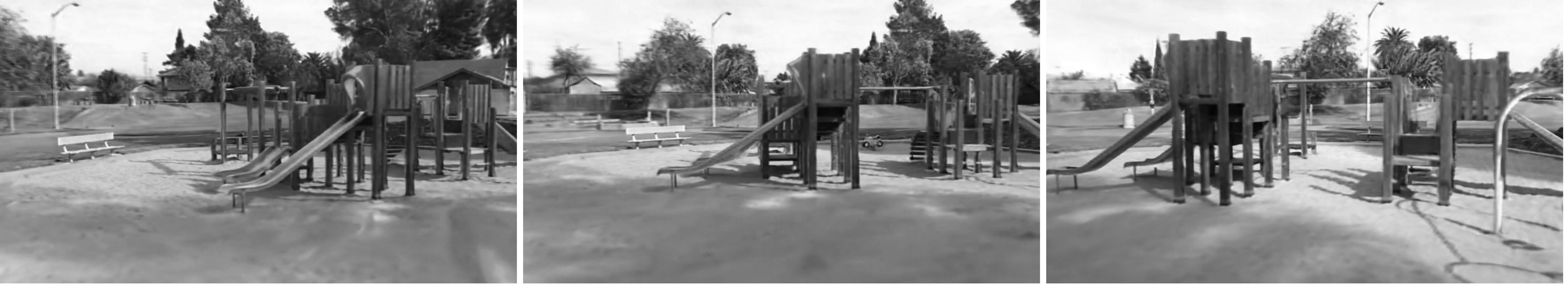}
    \caption{Novel views generated from the input grayscale images for \textit{playground} scene in Tanks \& Temples ~\cite{knapitsch2017tanks} dataset.}
    \vspace{-1mm}
    \label{fig:grey-scale-views}
\end{figure}

\begin{figure}[!t]
    \centering
    \includegraphics[width=\linewidth]{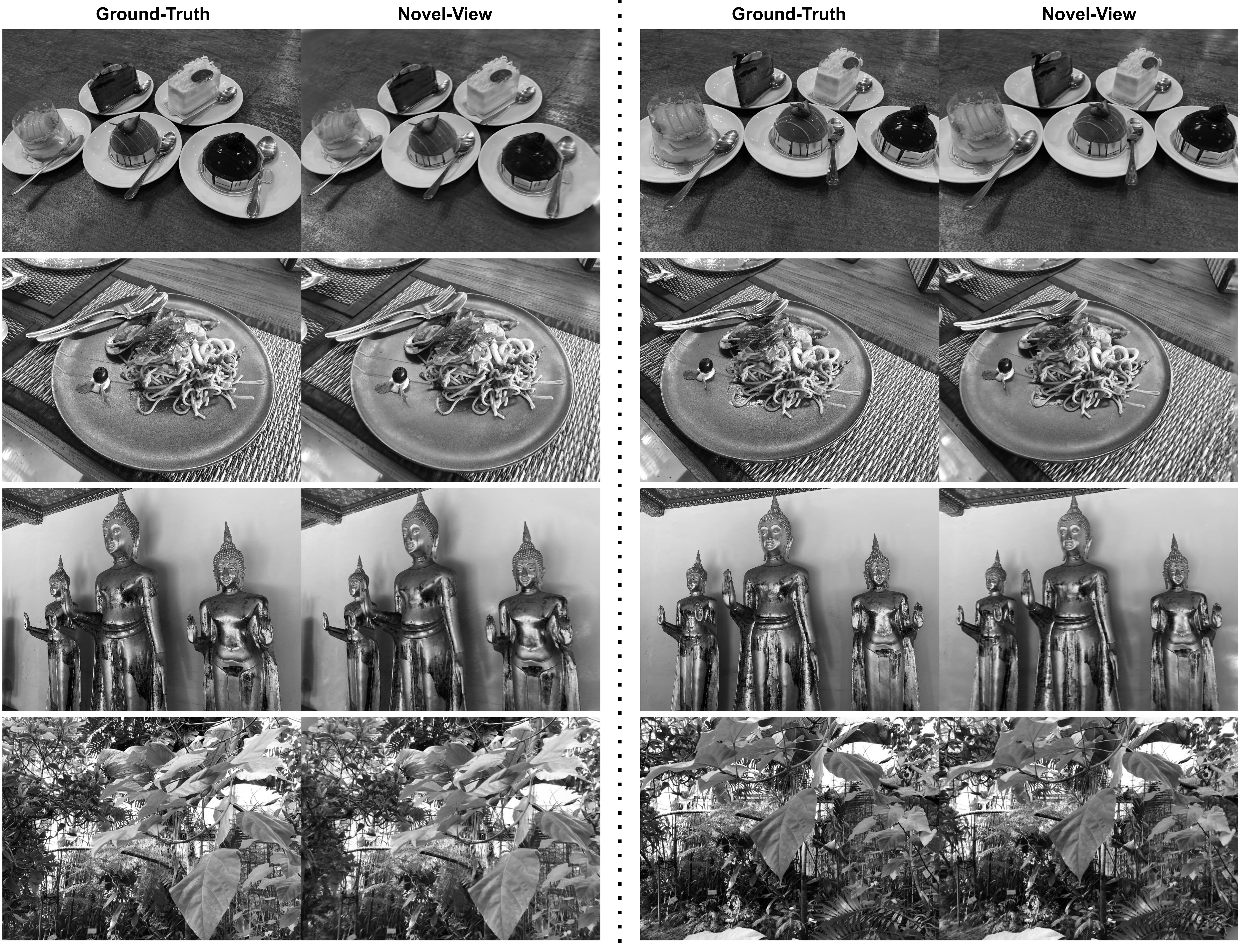}
    \caption{(Top to Bottom) : Comparison of ground-truth and novel-view for grayscale inputs for cake, pasta, buddha and leaves scene.}
    \label{fig:qual_grey_scale}
\end{figure}

We present quantitative results for generated grayscale novel views from ``Luma Radiance Field Stage'' (Stage 1) in Table ~\ref{tab:quant_grey_scale}. We also compare the generated novel-views with the ground-truth grayscale views in Fig.~\ref{fig:grey-scale-views} and~\ref{fig:qual_grey_scale}. We observe that generated novel-views are of good quality. This shows that learning monochromatic signal using a radiance field representation is achievable. 
\begin{table}[!t]
\caption{Quantitative analysis of GrayScale views}
\label{tab:quant_grey_scale}

\centering
\begin{tabular}{@{}ccccc@{}}
\toprule
               & \textbf{cake} & \textbf{pasta} & \textbf{buddha} & \textbf{leaves} \\ \midrule
\textbf{PSNR}  &        27.772       &     21.951           &       23.206          &       22.146          \\
\textbf{SSIM}  &       0.855        &        0.785        &         0.804        &        0.784         \\
\textbf{LPIPS} &       0.242        &      0.305          &       0.347          &        0.210         \\ \bottomrule
\end{tabular}
\end{table}

 \begin{figure}[!t]
    \centering
    \includegraphics[width=\linewidth]{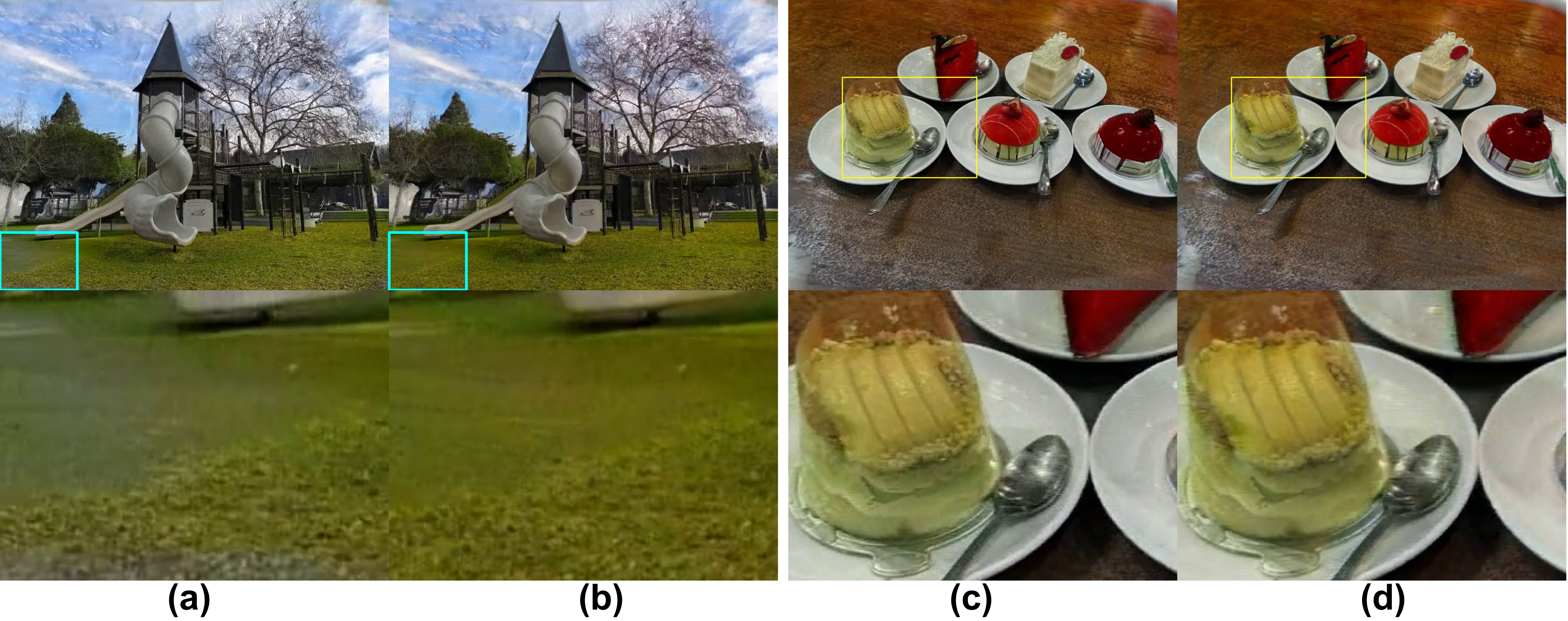}
    
    \caption{The effect of applying multi-scale regularization on the ``playground''( (a) and (b) ) and ``Cake'' ( (c) and (d)) scene. The highlighted region in the playground (b) and cake (d) had better color in the multi-scale regularization image (than the one w/o multi-scale regularization. Colors in w/o multi-scale regularization are slightly desaturated.}
    \label{fig:multi-scale-effect}
    \vspace{-1em}
\end{figure}

\begin{table}[t]
\caption{Characteristic comparison of Our method with Color-NeRF~\cite{cheng2024colorizing}}
\begin{adjustbox}{width=\linewidth}
\begin{tabular}{c|c|c|c}
\hline
Method    & Extra Parameters & Inference Speed & Supports other 3D representation \\ \hline
Color-Nerf~\cite{cheng2024colorizing} & Yes                    & High            & No                               \\
Ours      & No                     & Low             & Yes                             
\end{tabular}
\label{tab:colornerf_char}
\end{adjustbox}
\end{table}

\subsection{Impact of multi-scale regularization}
\label{sec:supp_ms_regularization}
We performed ablation studies on the impact of multi-scale regularization. When distilling color at the original resolution, some areas appeared de-saturated, as seen in the highlighted regions in Fig.~\ref{fig:multi-scale-effect} (a) \& (c). To overcome this issue, we employed multi-scale regularization, which mitigated the color de-saturation during the distillation process. This is evident in the improved color on the grass in playground and on top of the cake, as seen in Fig.~\ref{fig:multi-scale-effect} (b) \& (d). One can observe that a bluish patch is not there with the proposed multi-scale technique. These results demonstrate that our regularization method effectively addresses the color de-saturation problem in the generated views.

\subsection{Comparison with Color-NeRF~\cite{cheng2024colorizing}}
\label{sec:supp_comparison_colornerf}
\begin{figure}[t]
    \centering
    \includegraphics[width=\linewidth]{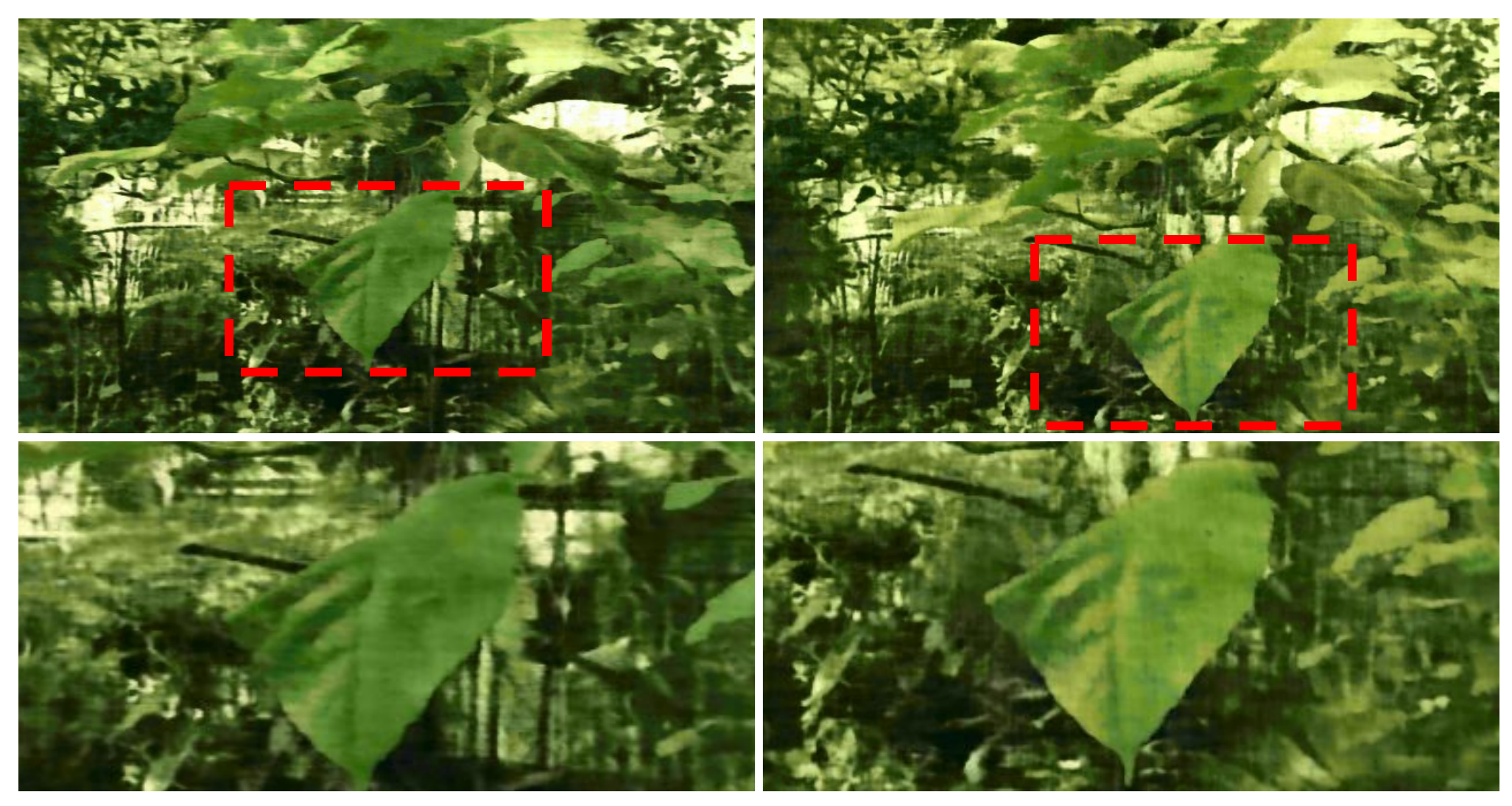}
    \caption{(Top row) Novel-views for ``leaves'' scene from ~\cite{cheng2024colorizing}. (Bottom row) Zoomed in region of the highlighted region. Notice the color change in the leaf.} 
    \label{fig:color_nerf_leaves}
\end{figure}
\begin{figure}[t]
    \centering
    \includegraphics[width=\linewidth]{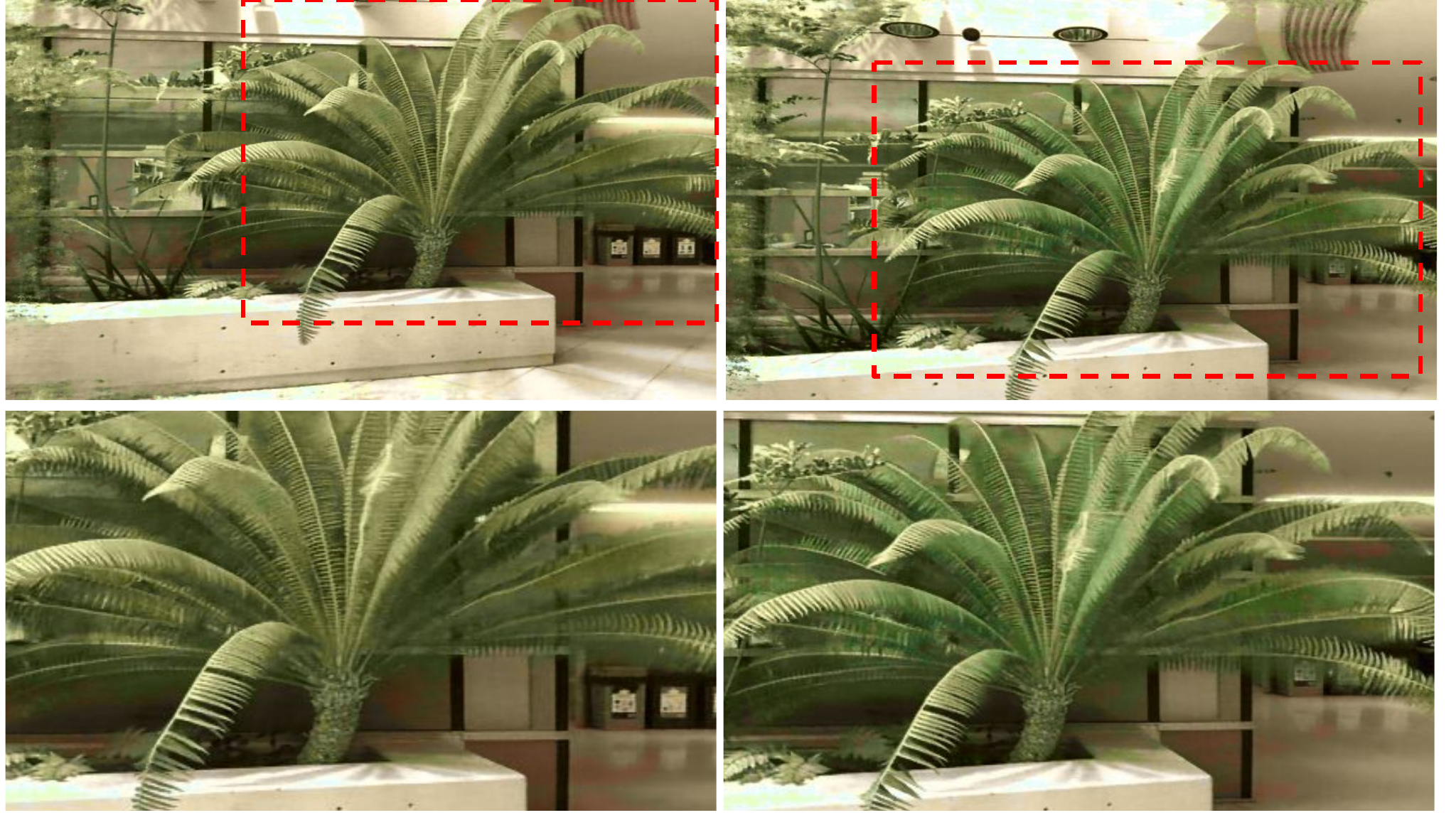}
    \caption{(Top row) Novel-views for ``fern'' scene from ~\cite{cheng2024colorizing}. (Bottom row) Zoomed in region of the highlighted region. Notice that the shade of the fern change from light green to a darker shade of green.} 
    \label{fig:color_nerf_fern}
\end{figure}
Color-NeRF~\cite{cheng2024colorizing} is a contemporary work that also solves a similar task. We show additional qualitative results from Color-NeRF in Fig.~\ref{fig:color_nerf_leaves} and~\ref{fig:color_nerf_fern}. We observe that cross-view consistency is not maintained by their method. Further, we compare with the cross-view consistency metrics described in the main paper. Tab.~\ref{tab:colornerf_quant} shows that our method performs better short-term and long-term consistency when compared with Color-NeRF. We also draw a comparison of their methodology with ours in Tab.~\ref{tab:colornerf_char}. We observe that whereas our method doesnot require any extra parameters to learn color. Color-NeRF requires a separate MLP to learn the color representation. Further, their method is too specific to NeRF architecture. Whereas ours can be easily used with any 3D representation.

\begin{table}[t]
\caption{Quantitative comparison of Our method with Color-NeRF~\cite{cheng2024colorizing}. Our method outperforms Color-NeRF for cross-view consistency.}
\begin{tabular}{@{}c|c|cc@{}}
\toprule
 &  & \textbf{pasta} & \textbf{fern} \\ \midrule
\begin{tabular}[c]{@{}c@{}}Short-Term\\ Consistency ($\downarrow$) \end{tabular} & Color-NeRF~\cite{cheng2024colorizing} & 0.077 & 0.021 \\
 & Ours & \textbf{0.009} & \textbf{0.010} \\ \midrule
\begin{tabular}[c]{@{}c@{}}Long-Term\\ Consistency ($\downarrow$)\end{tabular} & Color-NeRF~\cite{cheng2024colorizing} & 0.129 & 0.029 \\
 & Ours & \textbf{0.017} & \textbf{0.011} \\ \bottomrule
\end{tabular}
\label{tab:colornerf_quant}
\end{table}

\begin{figure}[t]
    \centering
    \includegraphics[width=\linewidth]{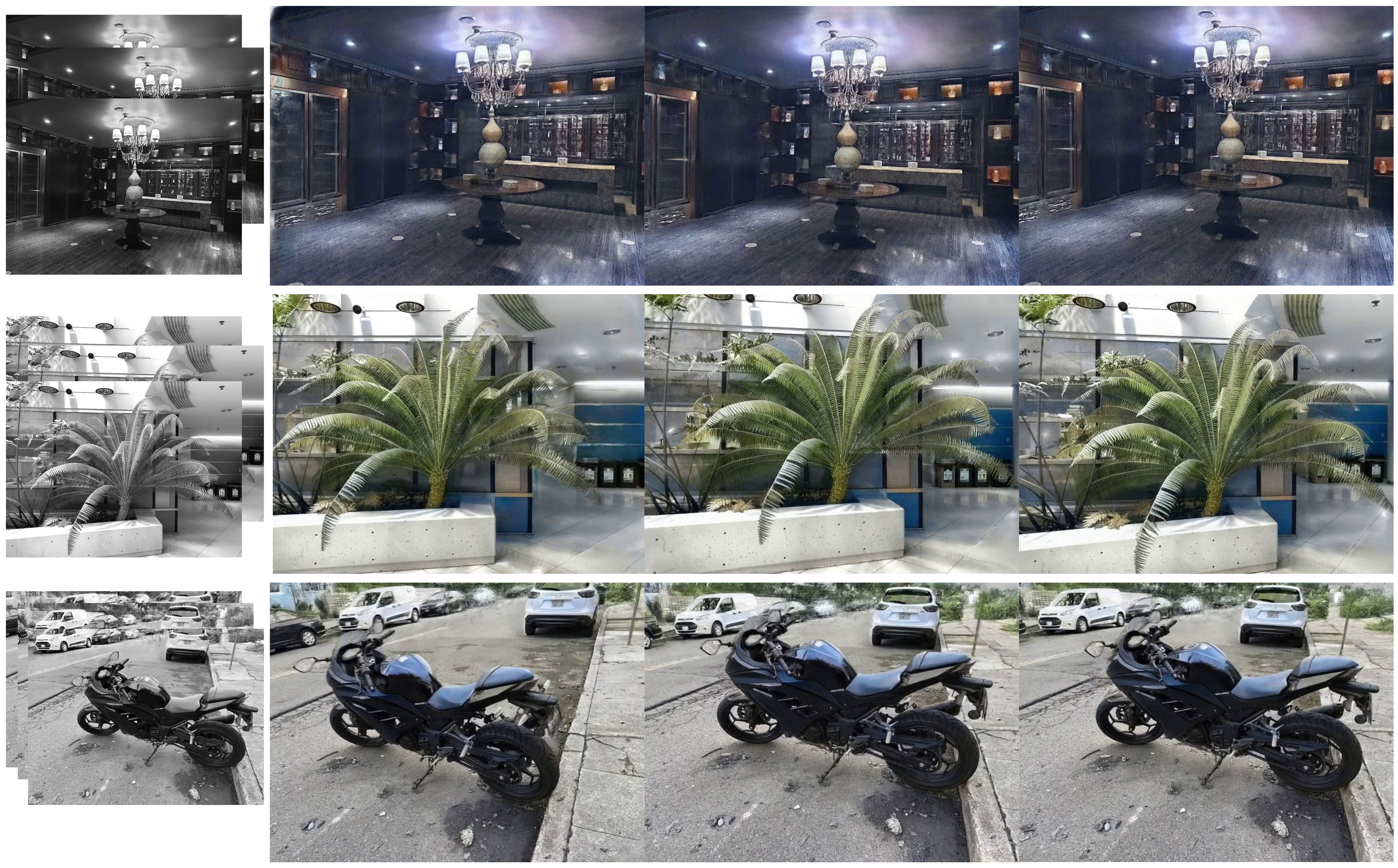}
    \caption{Novel views from the ``Different Room'', ``Fern'', and ``Ninja bike'' scenes are shown in the top, middle, and bottom rows, respectively. Note the consistency across views. To better appreciate these results, please refer to the supplementary video.}
    %\vspace{-10pt}
    \label{fig:novel_views}
\end{figure}

\subsection{Additional Results}
We present additional qualitative results in Fig. ~\ref{fig:novel_views}, Fig. ~\ref{fig:supp_1} and Fig. ~\ref{fig:supp_2}. We observe that our approach yields 3D consistent color views than the baseline methods. We also present quantitative results in Table ~\ref{tab:quant_supp_short_term} and ~\ref{tab:quant_supp_long_term}. Our method achieves better cross-view consistency compared with the baselines.

\begin{table}[t]
\vspace{-2mm}
\caption{Ablation results show that using the distillation strategy in the ``Lab'' color space leads to superior cross-view consistency performance across various scenes. }
\label{tab:ablation_color}
\begin{adjustbox}{width=\linewidth}
\begin{tabular}{c|cccc}
\toprule
  & \textbf{Cake} & \textbf{Pasta} & \textbf{Three Buddha} & \textbf{Leaves} \\ \midrule
\textbf{Ours(RGB)}         & 0.034 & 0.027 & 0.023 & 0.021 \\
\textbf{Ours(Lab)}        & \textbf{0.033} & \textbf{0.025} & \textbf{0.023} & \textbf{0.019}  \\ \bottomrule
\end{tabular}
\end{adjustbox}
\vspace{-15pt}
%\end{table}
%\end{minipage}
%\end{wrapfigure}
\end{table}
%\newpage

\subsection{Demonstration on Downstream task.}
\label{sec:suppl_downstream}
\begin{figure*}[!t]
    \centering
    \includegraphics[width=\linewidth]{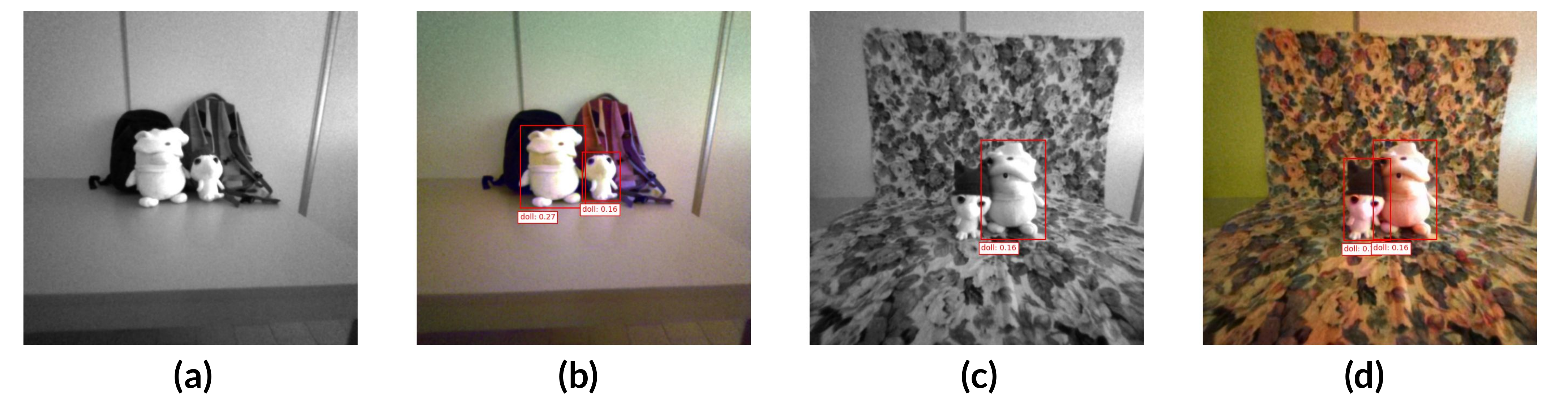}  

 \caption{To demonstrate the effectiveness of colorization, we conducted an object detection task on both original infrared (IR) views and their corresponding colorized counterparts. Notably, in (a), no objects are detected in the IR view, and only one out of two objects is detected in (c). However, objects are consistently detected in the colorized views, showcasing the enhanced performance achieved through colorization.}
%\vspace{-5pt}
\label{fig:ir_downstream}
\end{figure*}

We show downstream results in Fig.~\ref{fig:ir_downstream}. We observe that objects are consistently detected in the colorized novel-views. This downstream task is very useful to enable downstream tasks such as detection for IR sensors. 
\begin{figure}[t]
    \centering
    \includegraphics[width=\linewidth]{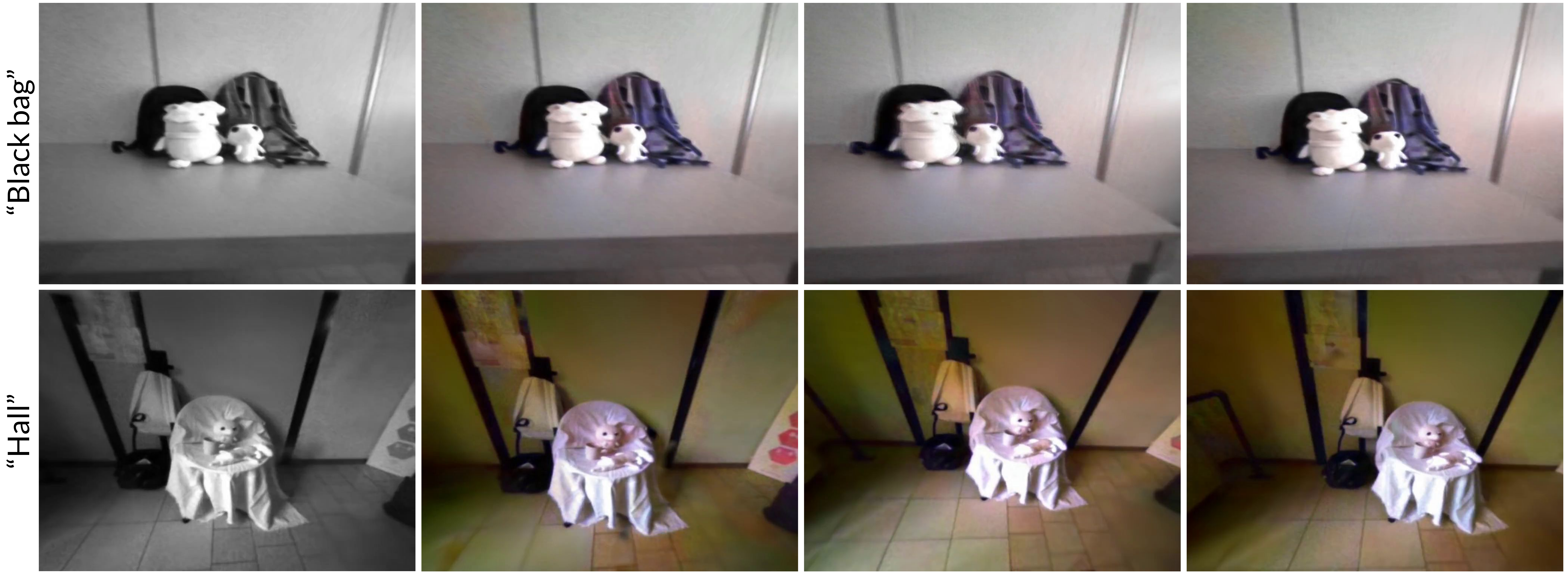}
    \caption{\textbf{More IR samples.} (Column 1) Input multi-view IR Sequence. (Columns 2, 3 \& 4) Colorized multi-views from Our method. Our approach yields consistent novel-views for a different input modality.}
    \label{fig:more_ir_samples}
\end{figure}

\subsection{Ablation on color-space}
We show ablation on color space in Tab.~\ref{tab:ablation_color}, We clearly see better cross-view consistency achieved with ``Lab'' color space.

\section{Discussion}
\subsection{Impact of Colorization Teacher Networks.}
The proposed method is compatible with any colorization technique. The quality of colorization depends on the selected teacher colorization network. By utilizing BigColor~\cite{kim2022bigcolor} and~\cite{zhang2016colorful}, our approach ensures multi-view consistency irrespective of the chosen teacher network. For instance, in the ``cake'' scene,~\cite{zhang2016colorful}. produce dull colors for various objects. In contrast, BigColor generates vivid and sharp colors for different objects. Likewise, we employ DDColor as the teacher colorization network in our infrared experiments.

\subsection{Video Colorization Baselines.} Video-colorization methods can generate different colorized outputs for differently rendered trajectories. For example, if we render N videos from N trajectories: ${T_1,T_2,....T_n}$ and feed them independently to a video colorization method, this can lead to different outputs even when the same reference images are given. Hence, even though feed-forward video colorization methods can generate temporally consistent views they do not guarantee 3D consistency. Compared to these baselines, our method ensures 3D consistency. 

\begin{table*}[t]
\caption{Quantitative results for short-term consistency}
\label{tab:quant_supp_short_term}
\begin{adjustbox}{width=\linewidth}
\begin{tabular}{@{}ccccc@{}}
\toprule
 \textbf{Scene}& \textbf{BigColor~\cite{kim2022bigcolor} $\rightarrow$ NeRF} & \textbf{NeRF $\rightarrow$ DeepRemaster~\cite{iizuka2019deepremaster}} & \textbf{NeRF $\rightarrow$ DeOldify~\cite{salmona2022deoldify}} & \textbf{Ours(BigColor ~\cite{kim2022bigcolor})} \\ \midrule
\textbf{pond} & 0.022 & 0.013 & 0.025 & \textbf{0.010} \\
\textbf{benchflower} & 0.025 & 0.013 & 0.022 & \textbf{0.010} \\
\textbf{chesstable} & 0.021 & 0.015 & 0.022 & \textbf{0.012} \\
\textbf{colorspout} & 0.025 & 0.013 & 0.031 & \textbf{0.011} \\
\textbf{lemontree} & 0.026 & 0.015 & 0.022 & \textbf{0.014} \\
\textbf{stove} & 0.014 & 0.010 & 0.019 & \textbf{0.008} \\
\textbf{piano} & 0.016 & 0.010 & 0.015 & \textbf{0.009} \\
\textbf{redplant} & 0.029 & 0.015 & 0.033 & \textbf{0.014} \\
\textbf{succulents} & 0.025 & 0.016 & 0.027 & \textbf{0.015} \\
\textbf{ninja} & 0.015 & 0.011 & 0.021 & \textbf{0.007} \\ \bottomrule      
\end{tabular}
\end{adjustbox}
\end{table*}

\begin{table*}[t]
\caption{Quantitative results for long-term consistency}
\label{tab:quant_supp_long_term}
\begin{adjustbox}{width=\linewidth}
\begin{tabular}{@{}ccccc@{}}
\toprule
 \textbf{Scene}& \textbf{BigColor~\cite{kim2022bigcolor} $\rightarrow$ NeRF} & \textbf{NeRF $\rightarrow$ DeepRemaster~\cite{iizuka2019deepremaster}} & \textbf{NeRF $\rightarrow$ DeOldify~\cite{salmona2022deoldify}} & \textbf{Ours(BigColor ~\cite{kim2022bigcolor})} \\ \midrule
\textbf{pond} & 0.035 & 0.017 & 0.028 & \textbf{0.015} \\
\textbf{benchflower} & 0.043 & 0.019 & 0.030 & \textbf{0.016} \\
\textbf{chesstable} & 0.033 & 0.023 & 0.028 & \textbf{0.021} \\
\textbf{colorspout} & 0.040 & \textbf{0.020} & 0.051 & \textbf{0.020} \\
\textbf{lemontree} & 0.041 & \textbf{0.020} & 0.027 & 0.021 \\
\textbf{stove} & 0.018 & 0.015 & 0.024 & \textbf{0.012} \\
\textbf{piano} & 0.026 & 0.014 & 0.019 & \textbf{0.013} \\
\textbf{redplant} & 0.041 & 0.021 & 0.041 & \textbf{0.020} \\
\textbf{succulents} & 0.040 & \textbf{0.024} & 0.032 & 0.026 \\
\textbf{ninja} & 0.021 & 0.015 & 0.027 & \textbf{0.012} \\\bottomrule      
\end{tabular}
\end{adjustbox}
\end{table*}

\begin{figure*}
    \centering
    \includegraphics[width=\linewidth]{Supp_Figures/Supp_Qualitative_1.pdf}
    \caption{\textbf{Qualitative results of our method with baselines.} We display two rows of each scene, each rendered from a different viewpoint. The first four columns depict the original resolution results, while the last four columns show zoomed-in regions of the highlighted areas in the first four columns. The baselines have color inconsistencies in their results, whereas our distillation strategy (columns  4 \& 8) maintains color consistency across different views. (Top to bottom) Order of scenes : pond, benchflower, chesstable, colorspout, lemontree}
    \label{fig:supp_1}
\end{figure*}
\begin{figure*}
    \centering
    \includegraphics[width=\linewidth]{Supp_Figures/Supp_Qualitative_2.pdf}
    \caption{\textbf{Qualitative results of our method with baselines.} We display two rows of each scene, each rendered from a different viewpoint. The first four columns depict the original resolution results, while the last four columns show zoomed-in regions of the highlighted areas in the first four columns. The baselines have color inconsistencies in their results, whereas our distillation strategy (columns  4 \& 8) maintains color consistency across different views. (Top to bottom) Order of scenes : stove, piano, redplant, succulents, ninja}
    \label{fig:supp_2}
\end{figure*}

\clearpage
%%%%%%%%% REFERENCES
{\small
\bibliographystyle{ieee_fullname}
\bibliography{main}
}

\end{document}